\documentclass[conference]{IEEEtran}
\IEEEoverridecommandlockouts
\usepackage[utf8]{inputenc}

\usepackage[
backend=biber,
style=ieee,
citestyle=numeric-comp,
sorting=none,
url=false,
isbn=false,
uniquename=false,
alldates=year,
clearlang=true,
minbibnames=8,
maxbibnames=10,
]{biblatex}

\DeclareFieldFormat[article,unpublished,misc,preprint]{title}{``#1''}

\let\oldcite\cite
\renewcommand{\cite}[1]{{\mbox{\oldcite{#1}}}}

\AtEveryBibitem{\clearlist{publisher}}
\AtEveryBibitem{\clearfield{note}}
\AtEveryBibitem{\clearlist{language}}
\AtEveryBibitem{\clearname{editor}}

\usepackage{doi}

\usepackage{booktabs}
\usepackage{siunitx}
\usepackage{hyperref}
\usepackage[font=small,labelfont=it]{caption}
\usepackage{graphicx}
\usepackage{subcaption}
\usepackage[dvipsnames, usenames]{xcolor}
\usepackage{float}
\usepackage{bm}

\let\rightarrow\veryshortarrow

\let\oldsubsection\subsection
\newcommand{\subsecno}[1]{\oldsubsection*{#1}\addcontentsline{toc}{subsection}{#1}}
\renewcommand{\subsection}[1]{\subsecno{#1}}

\usepackage{ragged2e}
\usepackage{placeins}
\usepackage{cancel}

\usepackage{cprotect}
\usepackage{amsmath,amsfonts,amssymb, bm}
\usepackage{lipsum}
\usepackage{acronym}
\usepackage{balance}

\acrodef{TDE}[TDE]{Time Difference Encoder}
\acrodef{DVS}[DVS]{Dynamic Vision Sensor}
\acrodef{CMOS}[CMOS]{Complementary Metal-Oxide-Semiconductor}
\acrodef{OF}[OF]{Optical Flow}
\acrodef{TDE}[TDE]{Time Difference Encoder}
\acrodef{EPSC}[EPSC]{Excitatory Postsynaptic Current}
\acrodef{LRS}[LRS]{Low Resistive State}
\acrodef{HRS}[HRS]{High Resistive State}
\acrodef{ITD}[ITD]{Interaural-Time-Difference}
\acrodef{ASE}[ASE]{Adaptive Synaptic Efficacy}
\acrodef{DPI}[DPI]{Differential Pair Integrator}
\acrodef{ASIC}[ASIC]{Application-Specific Integrated Circuit}
\acrodef{NAS}[NAS]{Neuromorphic Auditory Sensor}
\acrodef{DVS}[DVS]{Dynamic Vision Sensor}
\acrodef{ATIS}[ATIS]{Asynchronous time-based image sensor}
\acrodef{PCB}[PCB]{Printed Circuit Board}
\acrodef{AFM}[AFM]{Atomic Force Microscopy}
\acrodef{EOC}[EOC]{Electronic Oxide Cluster}
\acrodef{XPS}[XPS]{X-ray Photoelectron Spectroscopy}
\acrodef{AER}[AER]{Address-Event Representation}
\acrodef{FPGA}[FPGA]{Field-Programmable Gate Array}
\acrodef{XRD}[XRD]{X-ray Diffraction}
\acrodef{TEM}[TEM]{Transmission Electron Microscopy}
\acrodef{PCM}[PCM]{Phase Change Memories}
\acrodef{ECM}[ECM]{Electrochemical Metalization Memories}
\acrodef{VCM}[VCM]{Valence Change Memories}
\acrodef{PLD}[PLD]{Pulsed Laser Deposition}
\acrodef{XRR}[XRR]{X-ray Reflectivity}
\acrodef{IoT}[IoT]{Internet of Things}
\acrodef{MTE}[MTE]{Memristive Time Encoder}
\acrodef{FAC}[FAC]{Facilitatory}
\acrodef{TRG}[TRG]{Trigger}
\acrodef{LIF}[LIF]{Leaky Integrate and Fire}
\acrodef{DAC}[DAC]{Digital-to-Analog Converter}
\acrodef{ODE}[ODE]{Ordinary Differential Equation}
\acrodef{ARRE}[ARRE]{Average Relative Rotational Error}
\acrodef{IMU}[IMU]{Inertial Measurement Unit}
\acrodef{SLAM}[SLAM]{Simultaneous Localisation and Mapping}
\acrodef{VO}[VO]{Visual Odometry}
\acrodef{MOS}[MOS]{Metal Oxide Semiconductor}
\acrodef{PMOS}[pMOS]{p-type \ac{MOS}}
\acrodef{NMOS}[nMOS]{n-type \ac{MOS}}
\acrodef{MOSCAP}[MOSCAP]{\ac{MOS} Capacitor}
\acrodef{EXLIF}[ExLIF]{Exponential Leaky Integrate and Fire}
\acrodef{SMU}[SMU]{Source Measurement Unit}
\acrodef{GPU}[GPU]{Graphics Processing Unit}
\acrodef{DE}[DE]{Differential Equation}
\acrodef{ARRE}[ARRE]{Average Relative Rotational Error}
\acrodef{IC}[IC]{Integrated Circuit}
\acrodef{ISI}[ISI]{Interspike Interval}
\acrodef{VR}[VR]{Virtual Reality}
\acrodef{AR}[AR]{Augmented Reality}
\acrodef{CNN}[CNN]{Convolutional Neural Network}
\acrodef{SNN}[SNN]{Spiking Neural Network}
\acrodef{ANN}[ANN]{Artificial Neural Network}

\begin{document}

\title{Event-based vision for egomotion \\ estimation using precise event timing

\thanks{
\raggedright{
\hspace*{-1.33em}\textsuperscript{1} Bio-Inspired Circuits and Systems (BICS) Lab, Zernike Institute for Advanced Materials, University of Groningen, Netherlands.\\
\textsuperscript{2} Groningen Cognitive Systems and Materials Center (CogniGron), University of Groningen, Netherlands.\\
\textsuperscript{3} Neuronova Ltd., Italy. \\
\textsuperscript{4} Micro- and Nanoelectronic Systems (MNES), Technische Universit\"at Ilmenau, Germany. \\
\textsuperscript{5} Asynchronous VLSI and Architecture Group, School of Engineering \& Applied Science (SEAS), Yale University, CT, USA.\\
\textsuperscript{*} Corresponding author: h.r.greatorex@rug.nl
}}
}

\author{\IEEEauthorblockA{\textbf{Hugh Greatorex}\textsuperscript{1,2,*}, 
\textbf{Michele Mastella}\textsuperscript{3}, 
\textbf{Madison Cotteret}\textsuperscript{1,2,4}, \\
\textbf{Ole Richter}\textsuperscript{5},
\textbf{Elisabetta Chicca}\textsuperscript{1,2}
}
}

\maketitle
\begin{abstract}
Egomotion estimation is crucial for applications such as autonomous navigation and robotics, where accurate and real-time motion tracking is required. 
However, traditional methods relying on inertial sensors are highly sensitive to external conditions, and suffer from drifts leading to large inaccuracies over long distances.
Vision-based methods, particularly those utilising event-based vision sensors, provide an efficient alternative by capturing data only when changes are perceived in the scene. 
This approach minimises power consumption while delivering high-speed, low-latency feedback.
In this work, we propose a fully event-based pipeline for egomotion estimation that processes the event stream directly within the event-based domain. 
This method eliminates the need for frame-based intermediaries, allowing for low-latency and energy-efficient motion estimation.
We construct a shallow spiking neural network using a synaptic gating mechanism to convert precise event timing into bursts of spikes. 
These spikes encode local optical flow velocities, and the network provides an event-based readout of egomotion.
We evaluate the network's performance on a dedicated chip, demonstrating strong potential for low-latency, low-power motion estimation.
Additionally, simulations of larger networks show that the system achieves state-of-the-art accuracy in egomotion estimation tasks with event-based cameras, making it a promising solution for real-time, power-constrained robotics applications.
\end{abstract}

\begin{IEEEkeywords}
event-based vision, visual-odometry, neuromorphic computing, on-chip, spiking neural networks
\vspace{1em}
\end{IEEEkeywords}

\thispagestyle{plain}
\pagestyle{plain}

\begin{refsection}

\section{Introduction}

The estimation of egomotion plays an important role in applications such as autonomous navigation, robotics and \ac{AR}. 
Conventionally, vehicle egomotion is determined using a combination of wheeled odometry and inertial sensing~\cite{borenstein_mobile_1997}. 
However, this approach has its limitations: wheeled odometry is unreliable on slippery terrain, and inertial sensors are prone to drift over long distances, leading to inaccurate motion estimation~\cite{scaramuzza_visual_2011}.
In contrast, \ac{VO} has emerged as a robust and accurate alternative, offering improved relative positioning and trajectory estimation.
\ac{VO} also provides a complementary data source that can be fused~\cite{mueggler_continuous-time_2018} with \ac{IMU} data to maintain a reliable memory of an agent's location within its environment~\cite{nister_visual_2004, nister_visual_2006}.

Typically \ac{VO} relies on frame-based cameras, where the technology is mature and widely adopted.
However, these systems often fail to meet the demands of high-speed, low-latency applications~\cite{gehrig_low-latency_2024}.
High-speed cameras generate large data streams, are energy-intensive, and require bulky hardware, all of which impose significant constraints on real-time processing, especially when deployed on vehicle payloads.
To overcome these limitations, event-based vision sensors have emerged as an innovative alternative. 
Inspired by the biological retina, these sensors operate in an event-driven manner, capturing visual information with high temporal resolution and low-latency~\cite{mahowald1994,mead1988,lichtsteiner2008_fix}.
Their encoding of visual data, often as relative changes in luminance~\cite{posch2011,lichtsteiner2008_fix}, reduces power consumption and minimises redundant computation.
This makes them well-suited for applications requiring low-latency positional feedback, such as \ac{VO} on board autonomous drones~\cite{chen_event-based_2020, kaufmann_champion-level_2023}.
However, event-based cameras produce a continuous stream of asynchronous data (events), presenting emerging challenges for processing.
Much of the effort to address these challenges has been directed toward creating processing pipelines that match the accuracy of frame-based camera solutions.
A widely adopted approach involves converting event streams into frame-like representations by aggregating events over uniform time windows~\cite{zhu2018, nunes2022, lee2020, cuadrado2023, zhou_unsupervised_2017, censi2014, vitale2021, mitrokhin2018, mitrokhin_ev-imo_2019, gallego_event-based_2018, gallego_accurate_2017, renner_visual_2024}.
These event-frames are then processed using \ac{ANN} pipelines, such as \acp{CNN}, much like frame-based camera data.
However, this approach undermines the key advantages of event-based sensing, particularly its low-latency capability.
The conversion to frames inherently reintroduces latency, negating the original benefits of the event-driven paradigm.
To capitalise on the advantages of event-based sensors, it is crucial to adopt equally event-driven data processing methods~\cite{benosman_event-based_2014, haessig_spiking_2018, amir_low_2017}.
This remains an underexplored area of research.

\begin{figure*}
    \centering
    \includegraphics[width=\textwidth]{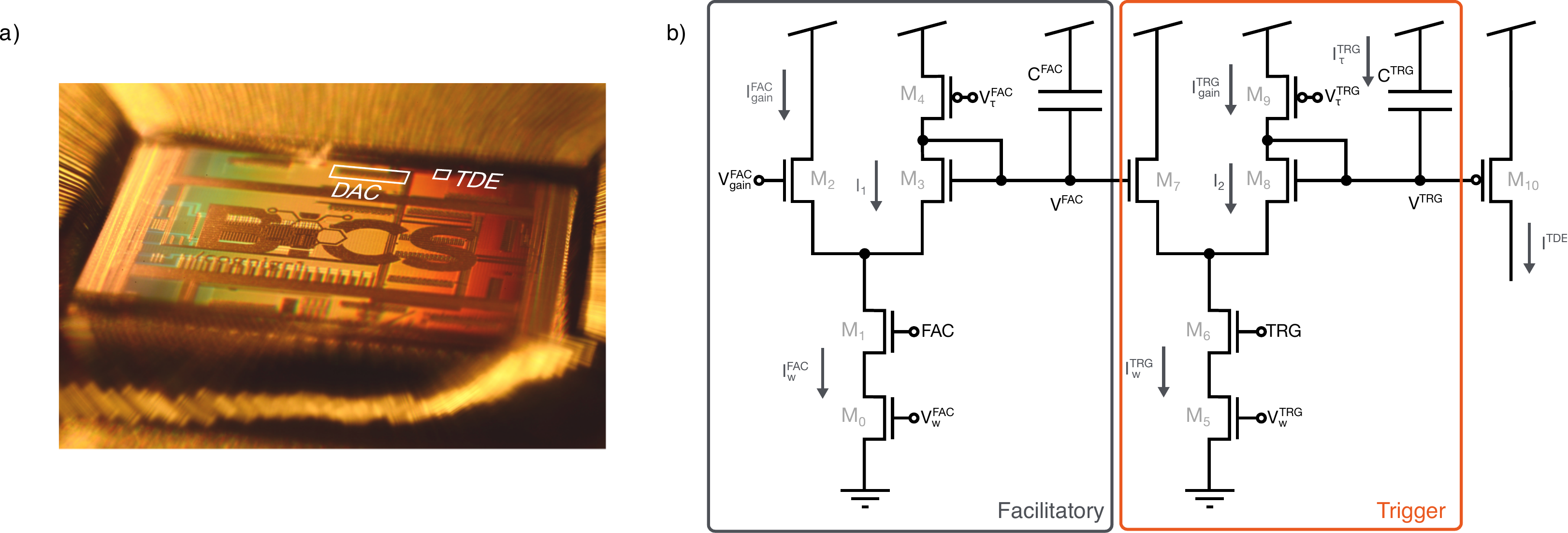}
    \vspace{2pt}
    \caption{\textbf{Photograph of the \textit{cognigr1} chip, fabricated in \qty[detect-all]{180}{\nano\meter} technology along with the schematic of the TDE synapse.} 
    \textbf{a)} The relevant structures on the die are indicated. 
    The total size of the TDE circuit is \qty{19}{\micro\meter}$\times$\qty{56}{\micro\meter} including guard rings and is biased by an on-chip DAC.
    \textbf{b)} Schematic of the state-of-the-art CMOS TDE synapse~\cite{greatorex_scalable_2025}, with facilitatory and trigger blocks labeled.}
    \label{fig:cognigr1_shem}
\end{figure*}

Recent advancements in neuromorphic computing have driven the development of dedicated event-based processors designed to handle the unique characteristics of event streams~\cite{yang_evgnn_2024, yao_spike-based_2024}.
These processors are optimised to preserve the high-speed, asynchronous nature of event readouts, making them well-suited for real-time or closed-loop applications.
However, many existing architectures remain limited in their ability to fully leverage the temporal precision of events for feature extraction. 
Most rely heavily on conventional models such as \ac{LIF} neurons, which, while widely used, are inadequate for capturing the richness of spike timing information in complex tasks, especially when embedded in shallow \acp{SNN}. 
This highlights the growing demand for processing units capable of efficiently and robustly extracting meaningful spatiotemporal features from high-throughput event streams.

In this work, we present a fully event-based pipeline for egomotion estimation that uses precise event timing to extract optical flow from the asynchronous output of event-based cameras.
Central to our approach is an on-chip implementation of the \ac{TDE}~\cite{milde2018, schoepe_finding_2024}, an event-based processing unit inspired by elementary motion detector circuits found in insect sensory pathways~\cite{borst2011, reichardt_autokorrelations-auswertung_1957}.
The \ac{TDE} converts the time differences between events on two input channels into an output event stream, capable of extracting spatiotemporal information from event-based data across various sensory domains~\cite{milde2018, dangelo_event-based_2020, schoepe2023, schoepe_finding_2024, schoepe_odour_2024, chiavazza_low-latency_2023}.
In the event-vision domain, a layer of \ac{TDE} elements is demonstrated in this work to provide a low-latency readout of egomotion.
To validate the proposed pipeline, we conducted two complementary studies.
We first applied an on-chip mixed-signal implementation of the \ac{TDE} model to an egomotion reconstruction task, using a  spatially-downsampled event stream captured by a car-mounted event camera in an urban environment. 
Then, we applied a scaled-up simulated network of \ac{TDE} elements to the full dataset.
The mixed-signal CMOS implementation demonstrates the viability of the approach for low-latency and energy-efficient operation, while the larger simulation highlights its scalability and suitability for real-world applications.

By avoiding frame-based intermediaries, our approach is well-suited for robotics tasks that demand real-time navigation on constrained power budgets. 
This advancement could enable more efficient and responsive autonomous systems, supporting future developments in low-latency, energy-constrained robotics~\cite{kaufmann_champion-level_2023,paredes-valles_fully_2024}.

\section{Results}

Here, we present the experimental validation and analysis of the \ac{TDE} circuit~\cite{greatorex_scalable_2025} (Fig.~\ref{fig:cognigr1_shem}), its application to an event-based egomotion estimation task when deployed in a lightweight \ac{SNN}, and the results of scaling up the \ac{SNN} in simulation. These simulations outperformed previous approaches in egomotion estimation accuracy across multiple dataset samples, demonstrating the scalability and robustness of the proposed network for real-world applications.

\begin{figure*}
    \centering
    \includegraphics[width=\linewidth]{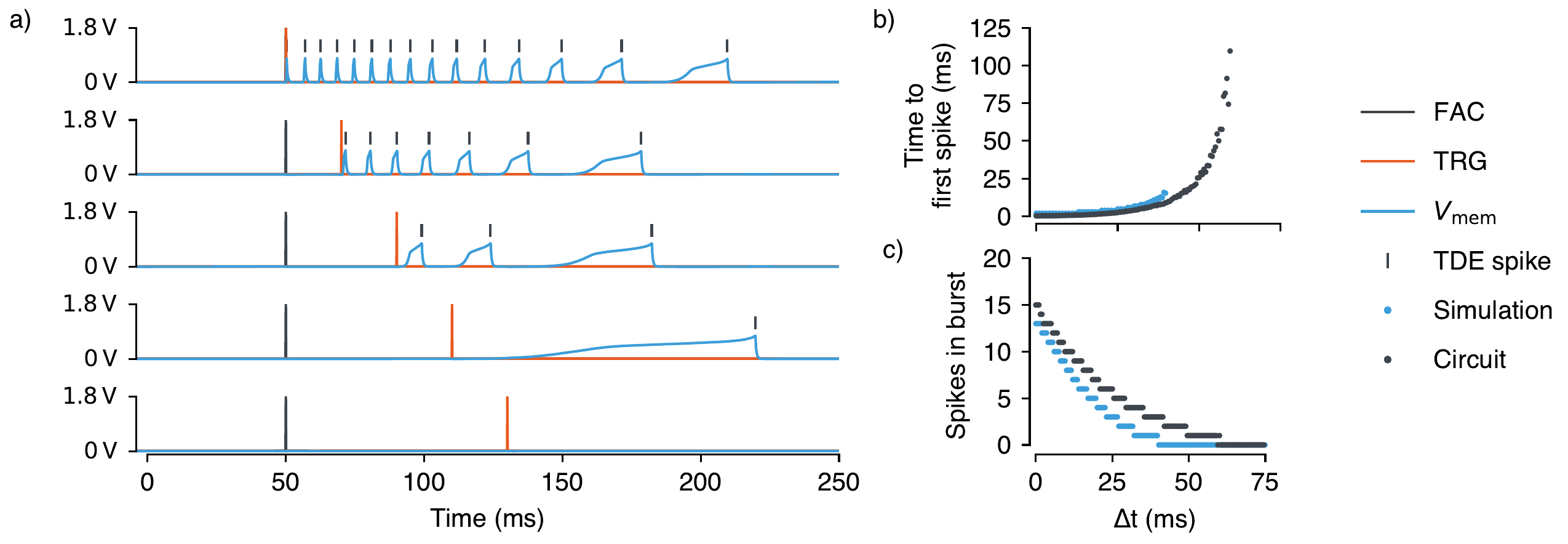}
    \caption{\textbf{Silicon measurements of the \ac{TDE} circuit.}
    \textbf{a)} The time ($\Delta t$) between FAC and TRG input events was increased systematically. Each plot shows a $\Delta t$ increment of \qty{20}{\milli \second} from \qty{0}{\milli \second} to \SI{80}{\milli \second}. 
    The membrane potential, $V_{\text{mem}}$ of the neuron integrates the input current from the \ac{TDE} synapse and outputs a spike when the threshold is reached.
    The encoding of the temporal distance between the input events is exhibited by the dynamics of $V_{\text{mem}}$ and the associated output spikes of the \ac{TDE}.
    \textbf{b)} The time to first spike of the \ac{TDE} spiking output for increasing time difference between input events. 
    The response of the simulated \ac{TDE} model to the same input is also shown as a comparison.
    \textbf{c)} The number of spikes in the burst of the \ac{TDE} with respect to the $\Delta t$ of input events. 
    }
    \label{fig:tde_circ_measurements}
\end{figure*}

\subsection{Time Difference Encoder Circuit}

The \ac{TDE} circuit~\cite{greatorex_scalable_2025} produces a subthreshold current \mbox{(\qty{}{\pico\ampere} - \qty{}{\nano\ampere})}, which inversely correlates with the time difference between two input events.
This current is then integrated by a silicon \ac{LIF} neuron~\cite{livi_current-mode_2009,chicca2014}, producing output spikes with an instantaneous frequency also inversely correlated with the input time difference.
This process utilises two distinct integrator blocks, called \textit{facilitatory} and \textit{trigger}, collectively termed the \ac{TDE} synapse, with each block receiving one of the two input channels of the \ac{TDE}.
Both integrator blocks initiate a time-decaying voltage trace in response to an event (or digital pulse).
Upon arrival of an event at the \ac{FAC} input, a decaying voltage trace is initiated. 
Concurrently, when an event arrives at the \ac{TRG} input, a synaptic current is generated, with its amplitude proportional to the instantaneous value of the trigger trace relative to the power supply.
This current is then integrated by a neuron circuit. 
The temporal proximity of \ac{FAC} and \ac{TRG} events determines the amplitude of the facilitatory trace at the arrival of the \ac{TRG} spike and the subsequent \ac{EPSC}.
As a result, the output spikes encode the input time difference through both their quantity and \ac{ISI}.
Thus, the \ac{TDE} circuit functions as an asymmetric correlation detector, encoding the time difference between input events in an analog manner.

The instantaneous magnitude of the current sourced by the \ac{TDE} synapse circuit, $I^{\text{TDE}}$, is proportional to the exponential of the time difference, $\Delta t_{\text{TDE}}$, between the \ac{FAC} and \ac{TRG} events.
This relationship is also influenced by the time constant of the facilitatory block and is described by the following equation: 
\begin{equation}
    I^\text{TDE} \propto  e^{\frac{-\Delta t_{\text{TDE}}}{\tau_\text{FAC}}}
    \label{eq:tde_current}
\end{equation}
The decay of this current, with this initial magnitude, is then described by:
\begin{equation}
    I^{\text{TDE}}(t) \propto  e^{-\frac{\Delta t_{\text{TDE}}}{\tau_\text{FAC}}}e^{-\frac{t}{\tau_{\text{TRG}}}}
    \label{eq:tde_current2}
\end{equation}
This confirms that the circuit's dynamics reflect the behaviour of the \ac{TDE} simulation model (Supplementary~\ref{sec:circuit_analysis}).
Specifically, there is an exponential relationship between $\Delta t_\text{TDE}$ and the initial magnitude of the current injected into the spiking neuron.

\subsection{TDE circuit measurements}


The response of the \ac{TDE} circuit to controlled input events was measured. 
In all experiments, the circuit was biased using the configurable \mbox{on-chip} \ac{DAC}. 
Fig.~\ref{fig:tde_circ_measurements}a shows the circuit's response to a sweep of input events during which the time difference, $\Delta t$, between \ac{FAC} and \ac{TRG} events was incremented from \qty{0}{} to \qty{80}{\milli\s}.
The membrane potential, $V_{\text{mem}}$, and spiking output of the \ac{TDE} are shown. 
It is evident that the circuit not only encodes the temporal separation of input events in the number of output spikes but also their relative timing within the burst.
\mbox{Fig.~\ref{fig:tde_circ_measurements}b} and \mbox{Fig.~\ref{fig:tde_circ_measurements}c} more explicitly plot this relationship, exemplifying how the \ac{TDE} circuit translates a temporal correlation into a burst of spikes (or events).
Additionally, the behaviour of the simulated \ac{TDE} model to the same input events is also shown, demonstrating the fidelity between the circuit and simulation model.
The parametrisation of the circuit and simulated model were not optimised for a specific task but chosen to have high spiking dynamic range over a $\Delta t$ of \qty{50}{\milli \second} (parameters listed in Supplementary Table~\ref{tab:circuit_parameters} and Table~\ref{tab:brian2_parameters}).

\begin{figure}[H]
    \centering
    \includegraphics[width=0.8\linewidth]{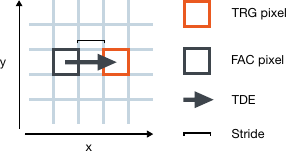}
    \caption{The \ac{TDE} can be applied to event-based vision tasks by ``connecting'' the \ac{FAC} and \ac{TRG} inputs to specific $(x,y)$ pixels.
    In this way the \ac{TDE} becomes receptive to a particular direction of motion. 
    In this case the \ac{TDE} is sensitive to left-right motion, indicated by the arrow representing the \ac{TDE} connectivity. 
    We define the stride as the pixel-wise separation of \ac{FAC} and \ac{TRG} inputs, in this example it is 1.}
    \label{fig:tde_pixels}
\end{figure}

\subsection{On-chip egomotion network emulation}

\begin{figure*}
    \centering
    \includegraphics[width=0.9\linewidth]{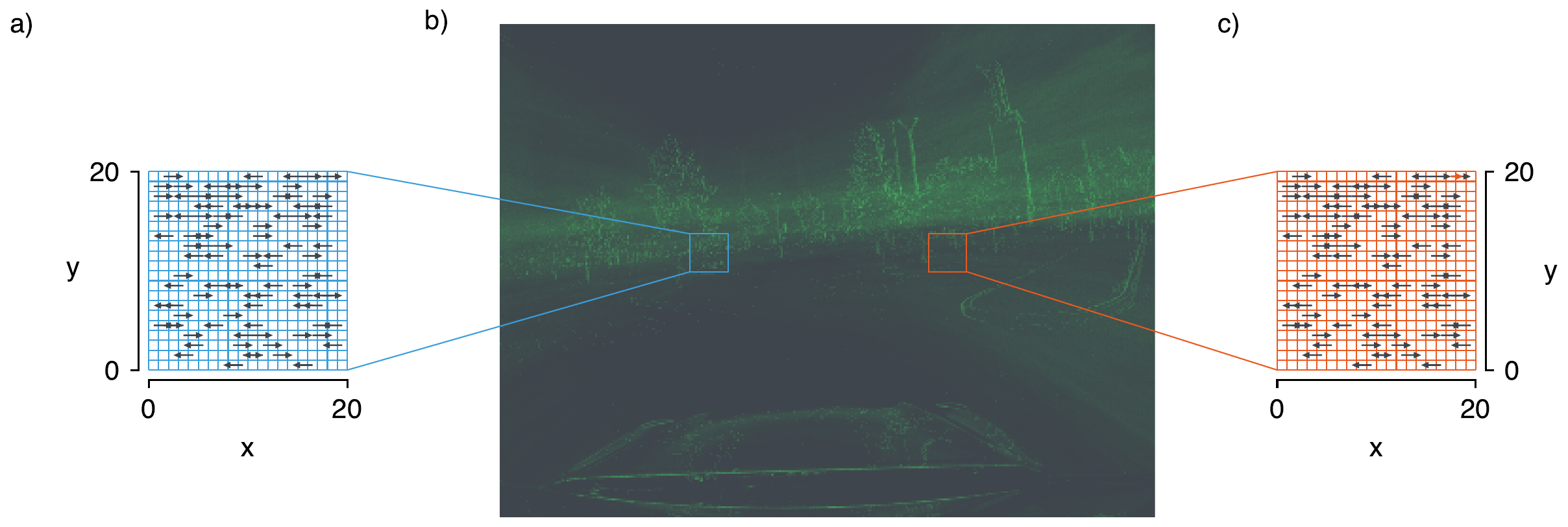}
    \caption{\textbf{An illustration of how the \ac{FAC} and \ac{TRG} connections are oriented for each \ac{TDE} unit in relation to the $\bm{(x, y)}$ event data from the event camera.}
    \textbf{b)} A single frame (generated by integrating activity over an arbitrary time step) of the MVSEC sample \textit{outdoor\_day1}. 
    This event data was recorded by a DAVIS 346B \ac{DVS} camera ($346 \times 260$ pixels) situated on the bonnet of a vehicle driving through a city. 
    The two boxes illustrate the two sample areas from which events were sampled to estimate the egomotion of the vehicle. 
    Both boxes, \textbf{a)} and \textbf{c)}, referred to as left and right, are $20 \times 20$ pixels in size and contain 100 randomly placed \acp{TDE}, with an equal proportion orientated with either left-right or right-left polarity and a stride of 2 pixels.
    Both sample boxes have precisely the same placement of \acp{TDE}.}
    \label{fig:on-chip_setup}
\end{figure*}

To apply the \ac{TDE} circuit to an event-based vision task, it is essential to establish some foundational intuition. 
The model is designed to process pairs of events through its two inputs. 
Conceptually, we can imagine ``connecting'' a single \ac{TDE} unit to two pixels of an event-based camera sensor, with each \ac{FAC} and \ac{TRG} input corresponding to a specific $(x,y)$ coordinate, or the events generated at that location.
When configured this way, the \ac{TDE} becomes sensitive to motion in a direction determined by the relative orientation of its \ac{FAC} and \ac{TRG} inputs within the $(x,y)$ coordinates of the sensor (Fig.~\ref{fig:tde_pixels}). 
By employing a population of \ac{TDE} units, it is possible to construct a network receptive to particular motion directions across the visual field.
Using the on-chip \ac{TDE} model, one can emulate such a network by sequentially providing the appropriate event-based data and measuring the circuit's response. 
As a proof of concept, this was able to be implemented with a single on-chip circuit, as the network operates in a purely feed-forward manner.

\begin{figure*}
\centering
\includegraphics[width=0.9\textwidth]{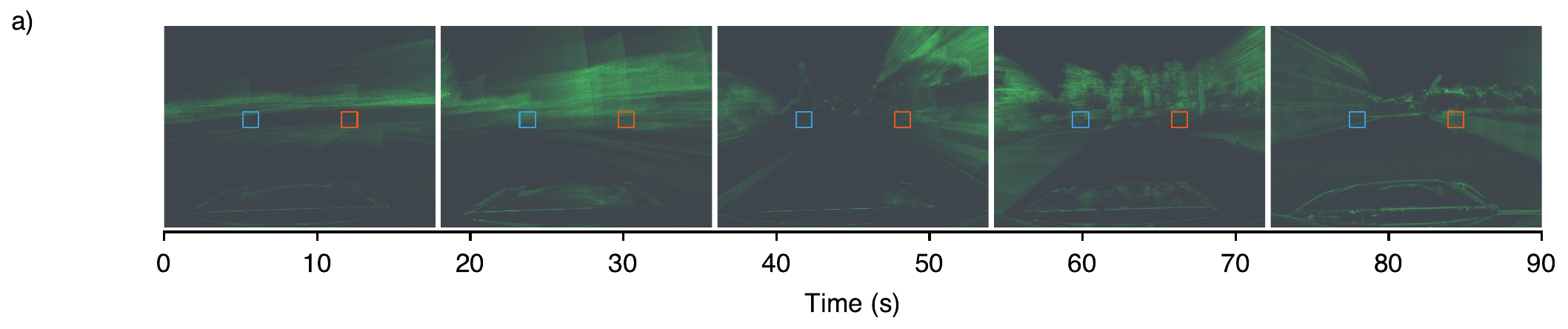}
\includegraphics[width=0.9\textwidth]{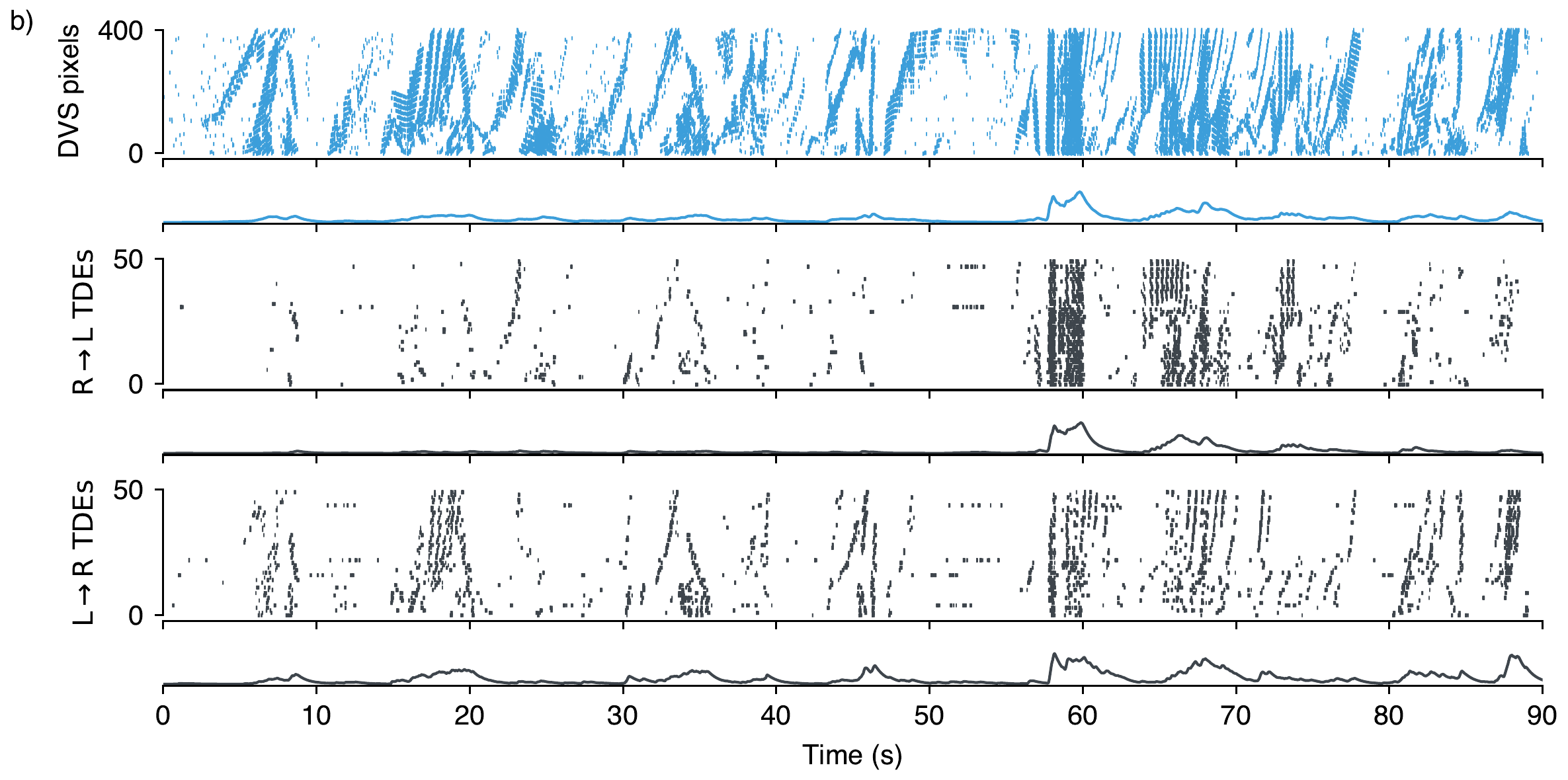}
\includegraphics[width=0.9\textwidth]{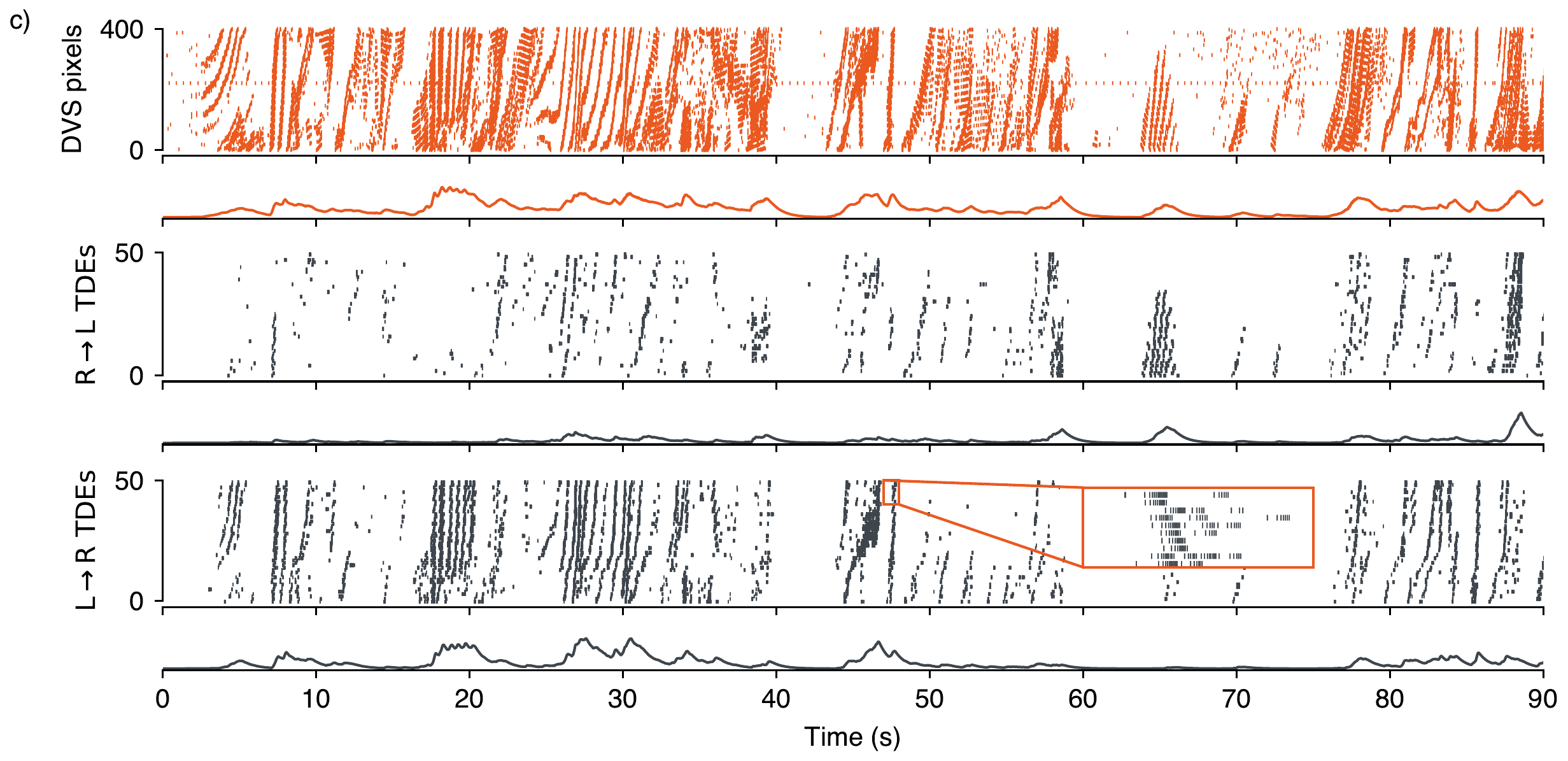}
\caption{\textbf{Measurements from the \ac{TDE} circuit on the \textit{cognigr1} chip, implementing the egomotion estimation task.}
For this task the first 90 seconds of events from the \textit{outdoor\_day1} MVSEC sample were used.
\textbf{a)} The event data from the event-camera mounted on the driving car. 
The two boxes (each $20\times20$ pixels) depict the two areas of the visual field from which events were sampled.
\textbf{b)} The input events from the left box (blue) and the response of \acp{TDE} in the network orientated in the right-left ($\text{R} \rightarrow \text{L}$) and left-right ($\text{L} \rightarrow \text{R}$) direction. 
Below each raster plot the integrated activity, normalised for each box, is shown. 
This activity shows the differing sensitivity of \ac{TDE} orientation and the event rate propagated through the network.
\textbf{c)} The same information for the right box (orange). 
Additionally, a zoomed in section of the \ac{TDE} network response is displayed to show the bursting activity of individual \ac{TDE} units.}
\label{fig:boxIO}
\end{figure*}

\begin{figure*}
    \centering
    \includegraphics[width=\textwidth]{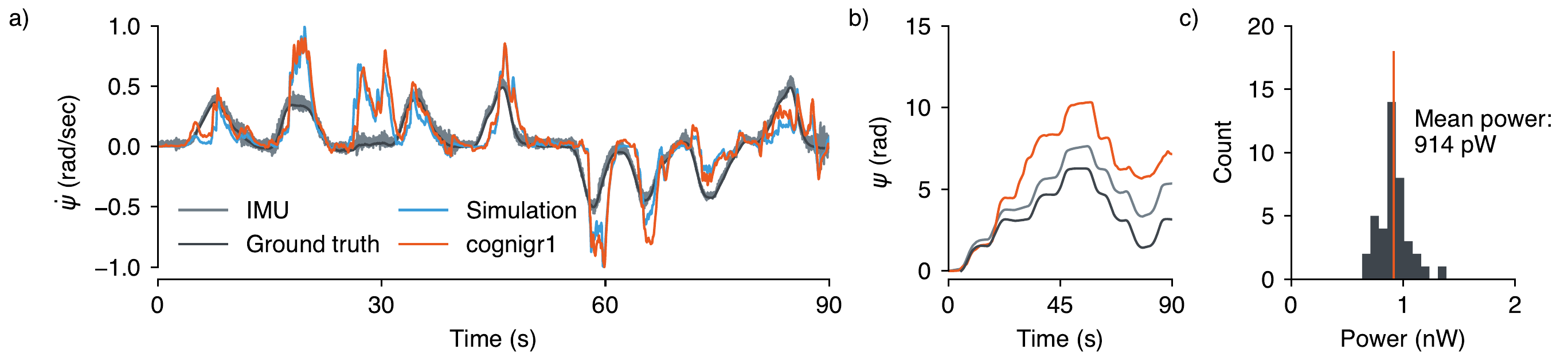}
    \caption{\textbf{The egomotion estimation of the vehicle using the \textit{congigr1} chip.}
    \textbf{a)} The egomotion network activity compared with the \ac{IMU} measurements from the vehicle and the recorded ground truth. The estimated motion is the angular velocity of the car, $\dot{\psi}$, as it turns, otherwise known as the yaw. 
    The network activity plotted is from both on-chip measurements and \textit{Brian2} simulations with exactly the same connectivity and input events. 
    An \ac{ARRE} of \qty{0.00014}{\radian} is reported for both the simulated and on-chip network over the duration of the sample.
    \textbf{b)} The accumulated heading direction of the vehicle, $\psi$, constructed by integrating $\dot{\psi}$.
    \textbf{c)} On-chip power measurements of the \ac{TDE} circuitry while performing the egomotion task, the average power consumption was measured for 40 random \ac{TDE} instances within the emulated network. 
    An average power consumption of \qty{914}{\pico \watt} is reported per \ac{TDE} in the network.}
    \label{fig:chip_results}
\end{figure*}

For this task, an \ac{SNN} comprising of 200 \ac{TDE} units was used. 
The \acp{TDE} were distributed within the visual field, as depicted in Fig.~\ref{fig:on-chip_setup} and were positioned to be sensitive to the vehicle's yaw rate, or, in other words, its turning velocity.
Fig.~\ref{fig:boxIO} shows the IO events of the network emulated on-chip.
Input data from the MVSEC~\cite{zhu2018} sample \textit{outdoor\_day1} is shown for both boxes positioned within the visual field \mbox{(Fig.~\ref{fig:on-chip_setup})}, along with the spiking response of the \ac{TDE} populations.
The estimated egomotion signal was generated by integrating the \ac{TDE} spiking activity and computing the difference between the right-left and left-right oriented \acp{TDE}. 
This process was applied to both boxes, and the activity differences for each box were summed.
The resulting signal was then normalised to the maximum absolute value over the duration of the sample.
Fig.~\ref{fig:chip_results} displays the output of the on-chip network, the \ac{IMU} recording and vehicle ground truth. 
The ground truth was generated using a fusion of LiDAR, \ac{IMU}, motion capture, and GPS data~\cite{zhu2018}.
The egomotion signal representing the yaw rate of the vehicle ($\dot{\psi}$) was measured with an \ac{ARRE} (Supplementary~\ref{sec:arre}) of \qty{0.00014}{\radian} on-chip.
As a comparison, the same network configuration with the same input events was simulated using the \ac{SNN} simulator \textit{Brian2}, demonstrating high similarity between the \ac{TDE} simulation model and its circuit implementation.

\subsection{On-chip network power consumption}

We measured the power consumption of the \ac{TDE} circuit while sequentially executing the network.
A histogram of the average power consumption per \ac{TDE} over the duration of the event-camera sample is shown in \mbox{Fig.~\ref{fig:chip_results}c}. 
A mean power consumption per \ac{TDE} of \qty{914}{\pico\watt} was measured.
Therefore, when scaled by the number of units used in the network emulated on-chip \mbox{(Fig.~\ref{fig:boxIO})}, a total power consumption of \qty{1.8}{\nano\watt} is estimated for the \acp{TDE} in the network when performing the task.
However, it is important to note that if such a system were fully realised in hardware, there would be additional power overhead associated with aggregating spiking activity to provide a readout of egomotion, as this process was performed off-chip in this work.

\subsection{Scaled-up simulations}

\begin{figure*}
    \centering
    \includegraphics[width=\linewidth]{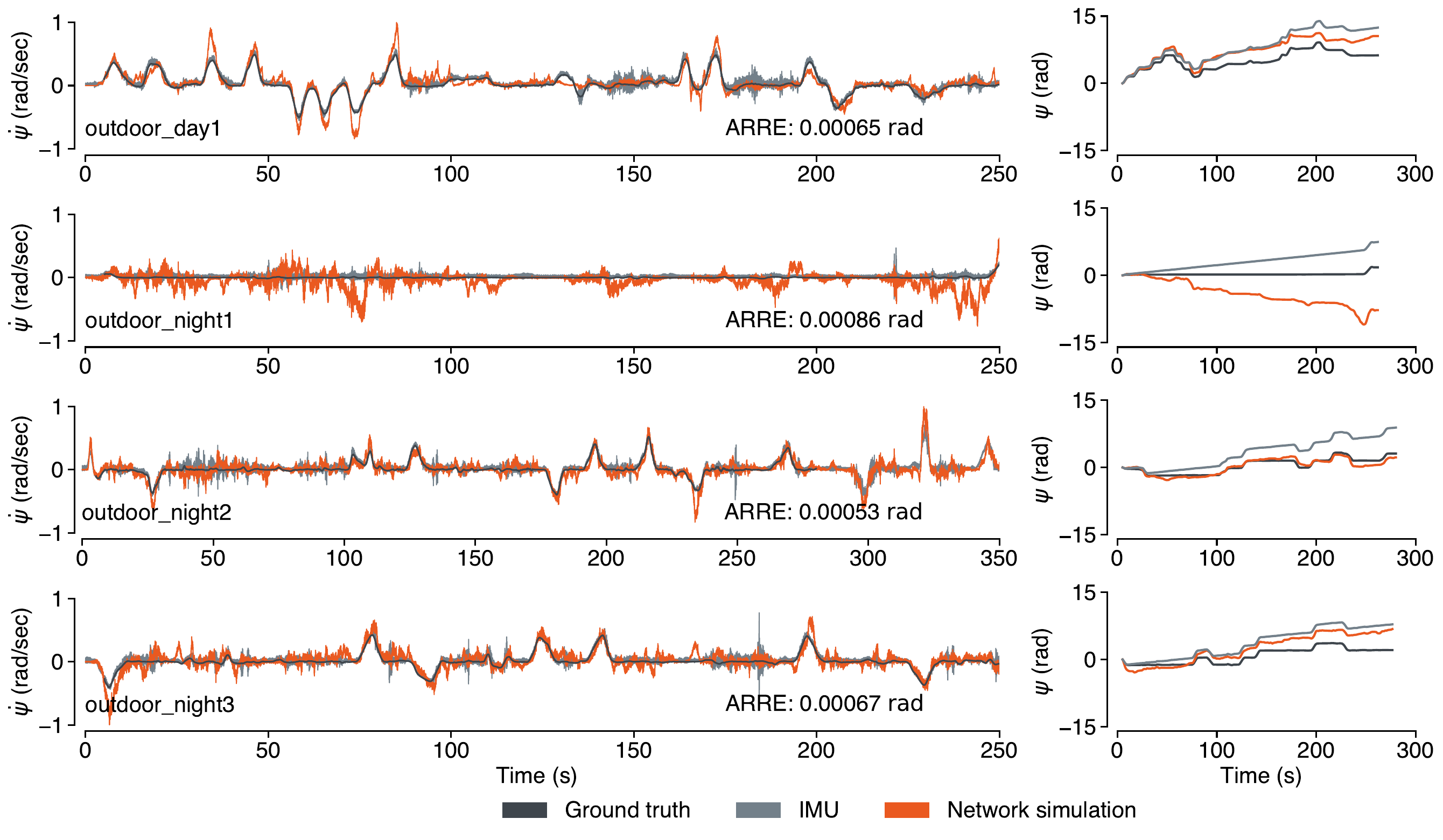}
    \caption{\textbf{The simulated scaled-up egomotion network sampling from the entire visual field.}
    The network estimation is compared to the \ac{IMU} recording and ground truth.
    The \ac{ARRE} compared to the ground truth is reported for each driving sequence and the accumulated heading direction, $\psi$, is compared to the ground truth value and the value derived from the on-board \ac{IMU}.}
    \label{fig:tde_mvsec_network_full}
\end{figure*}

To assess the full-scale performance of the egomotion network, its size was increased.
With the limitation of only a single \ac{TDE} circuit on the \textit{cognigr1} die these networks were run in simulation using the \ac{TDE} model.
While the network emulated on-chip was limited to receiving sparse input from two locations in the visual field, the full-scale network densely sampled the entire visual field, amounting to 178,880 \ac{TDE} units in total (compared to 200 in the network emulated on-chip).
This is comparable to the quantity of spiking neurons available on several cores of state-of-the-art neuromorphic processors.~\cite{orchard_efficient_2021}.
An identical approach of integrating the activity of left-right and right-left orientated \ac{TDE} units and evaluating their difference in activity was applied.
The results of these simulations are shown in \mbox{Fig.~\ref{fig:tde_mvsec_network_full}}, again with comparison against the \ac{IMU} and ground truth.
It is noted that in three cases (\textit{outdoor\_day1}, \textit{outdoor\_night2}, and \textit{outdoor\_night3}) the accumulated heading direction, $\psi$, suffers from less drift than the signal derived from the \ac{IMU} when compared to the ground truth.
The \ac{ARRE} of the egomotion estimation for each sample, shown in \mbox{Table~\ref{tab:arre}}, highlights an improvement of at least one order of magnitude over previous works using the same dataset. 
The proposed network achieves consistently lower error across all samples, demonstrating its effectiveness in enhancing egomotion estimation accuracy.

\begin{table*}
    \centering
    \begin{tabular}{l*{5}{c}}
    \toprule
    & \multicolumn{4}{c}{\ac{ARRE} (\qty{}{\radian})}\\
    Comparison & outdoor\_day1 & outdoor\_night1 & outdoor\_night2 & outdoor\_night3 \\
    \midrule
    \textbf{This work} & \textbf{0.00065} & \textbf{0.00086} & \textbf{0.00053} & \textbf{0.00067} \\ 
    \cite{mitrokhin2019} & 0.0994 & 0.0632 & 0.116 & 0.121 \\
    \cite{ye20} (SfMLearner~\cite{zhou_unsupervised_2017}) & 0.00916 & 0.00433 & 0.00499 & 0.00482 \\
    \cite{ye20} & 0.00267 & 0.00139 & 0.00202 & 0.00202 \\
    \cite{zhu_unsupervised_2019} & 0.00867 & - & - & - \\
    \bottomrule
    \end{tabular}
    \caption{\textbf{Comparisons of the egomotion network in this work with other works estimating egomotion using the same dataset.}}
    \label{tab:arre}
\end{table*}

\section{Discussion}

In this work, we leveraged event-based vision and processing within the event-based domain to enable accurate egomotion estimation, optimised for ultra-low power operation.
Unlike current methods that convert event streams into frames for processing~\cite{mitrokhin2019, zhou_unsupervised_2017, zhu_unsupervised_2019, ye20}, our method remains entirely event-based, enabling a significant reduction in latency and power consumption when executed on dedicated neuromorphic hardware.
Notably, self-motion reconstruction in this work outperformed comparable event-framing-based approaches (Table~\ref{tab:arre}).
A key strength of our approach lies in its simplicity and generalizability; the proposed network architecture does not rely on parameter optimisation or learning~\cite{gehrig_e-raft_2021,paredes-valles_taming_2023, shiba_secrets_2024}, making it robust to diverse and novel visual inputs.
This contrasts with learning-based methods, especially those using \acp{CNN}, which often encounter difficulties in transferring across different domains and datasets. 
Such approaches are also particularly fragile when processing scenes consisting of independently moving objects.
In this context, the configuration of the \ac{TDE} units' receptive fields, spatial distributions, and parameterisation required minimal engineering, leaving ample room for future exploration and optimisation. 
Despite the lack of optimisation, the results of our lightweight and shallow network architecture highlight the inherent effectiveness of our approach.

The network was validated through emulation on a custom mixed-signal subthreshold \ac{IC}, achieving competitive results against both \ac{IMU}-based reconstructions and state-of-the-art event-based \ac{VO} techniques.
Despite sampling less than 1\% of the visual field, the on-chip method accurately reconstructed self-motion all while consuming an estimated \qty{1.8}{\nano\watt}. 
When scaled to the full visual field, simulations suggest that a hardware implementation of a \ac{TDE} network on-chip could operate at a power budget of approximately \qty{200}{\micro\watt}, making it highly suitable for power-constrained applications such as micro-drones, edge computing devices, and \ac{VR} headsets~\cite{palossi_64-mw_2019}.
A significant advantage of the proposed pipeline is its resource-efficient structure, as it eliminates the need for pre-processing steps like hot-pixel removal or spatiotemporal down-sampling~\cite{yedutenko_tde-3_2024, dangelo_event-based_2020}.
The potential for direct integration with event-based sensors would streamline the pipeline, aligning with advancements in custom asynchronous \acp{ASIC} designed for event-driven architectures~\cite{yao_spike-based_2024, lefebvre_mixed-signal_2024}.
Such integration further reinforces the feasibility of real-time, fully event-based processing pipelines in robotics systems~\cite{schoepe_finding_2024,paredes-valles_fully_2024}.

A comparative analysis with \ac{IMU}-based heading direction systems revealed instances where our event-based network outperformed the \ac{IMU}, particularly in scenarios where drift impacted inertial data. 
This suggests that our approach could augment \ac{IMU}-based systems by recalibrating allocentric maps in real-time using visual inputs, particularly in GPS-denied environments. 
The capacity to leverage visual information in such scenarios highlights the broader utility of event-based egomotion estimation for maintaining robust and accurate navigation.
Finally, the low-power consumption and efficient spike-timing-based processing of our approach make it particularly advantageous for platforms that require low-latency feedback on stringent energy constraints. 
Applications such as \ac{VR} headsets or micro-drones, where power budgets are heavily limited, could benefit from the sparsification of computation enabled by this method. 
By addressing the dual challenges of latency and energy efficiency, this work offers a compelling alternative to traditional approaches, offering a promising solution for next-generation, low-power, real-time \ac{VO} systems.

\section{Methods}
\subsection{TDE circuit working principle}

The first circuit implementation of the \ac{TDE} model was previously proposed and tested in~\cite{milde2018}. 
In this work, we introduce a modified version of the synaptic circuit~\cite{greatorex_scalable_2025} that employs two \ac{DPI} circuits~\cite{bartolozzi2007} to integrate \ac{FAC} and \ac{TRG} inputs on their respective tails. 
Similar to the algorithmic model, the CMOS implementation consists of a circuit for both the \ac{TDE} synapse and the spiking neuron.
The schematic for the \ac{TDE} synapse is shown in \mbox{Fig.~\ref{fig:cognigr1_shem}b} and the neuron schematic is shown in Supplementary \mbox{Fig.~\ref{fig:neuron}}, both circuits were realised using \ac{MOS} technology and operate in the subthreshold regime.

When an event (digital pulse) arrives at the \ac{FAC} transistor, the tail of the first \ac{DPI} drains charge from the capacitor $C^{\text{FAC}}$ (fabricated as a \ac{MOSCAP}).
The magnitude of discharge is determined by the biases $V^\text{FAC}_\text{w}$ and $V^{\text{FAC}}_\text{gain}$ (Supplementary~\ref{sec:circuit_analysis}). 
The capacitor $C^{\text{FAC}}$ then recharges at a rate determined by the bias $V^{\text{FAC}}_\tau$, thus raising the voltage $V^{\text{FAC}}$ to $V_{\text{dd}}$ with a time constant, $\tau_{\text{FAC}}$.
If, within this facilitation time, a \ac{TRG} event arrives at the tail of the second \ac{DPI} the voltage on the gate of transistor $\text{M}_{7}$ is sufficient to elicit the discharge of capacitor $C^{\text{TRG}}$.
This magnitude of this discharge is determined not only by the time since \ac{FAC} events but also by the bias $V^{\text{TRG}}_\text{w}$. 
In a similar way, the \ac{TRG} capacitor recharges at a rate determined by the bias $V^{\text{TRG}}_\tau$, thus whilst the voltage $V^{\text{TDE}}$ is below $V_{\text{dd}}$ a current is conveyed through \ac{PMOS} transistor $\text{M}_{10}$.
The \ac{LIF} circuit that integrates this current, $I^{\text{TDE}}$, elicits spiking activity, provided the current dynamics and neuron biasing are adequate.
The neuron circuit fabricated in this work was derived from the \ac{DPI} neuron first introduced in~\cite{chicca2014} and its behaviour is explained more comprehensively in Supplementary \mbox{Fig.~\ref{fig:neuron}}.

The aforementioned circuits were fabricated using the \qty{180}{\nano \meter} X-FAB process on the \textit{cognigr1} \ac{ASIC}. The chip includes a single \ac{TDE} and an on-chip 12-bit \ac{DAC}, which supplies the subthreshold voltage parameters necessary for biasing the circuitry. To enable power measurements, the power supply of the \ac{TDE} circuitry is isolated. A photograph of the fabricated chip is provided in \mbox{Fig.~\ref{fig:cognigr1_shem}a}, highlighting the approximate location of the circuits on the die.

\subsection{TDE simulation model}

The simulated \ac{TDE} model has two inputs: \acf{FAC} and \acf{TRG}, both of which receive events.
The FAC input events are integrated onto the trace $I_\text{FAC}$, with dynamics given by
\begin{equation}
    \frac{\text{d}I_{\text{FAC}}}{\text{d} t} = - \frac{I_{\text{FAC}}}{\tau_{\text{FAC}} } + \sum_{i} w^{\text{FAC}} \delta(t - t^\text{FAC}_{i})
\label{eqn:fac}
\end{equation}
where $\tau_{\text{FAC}}$ is the time constant of the decay, $\delta(t - t^\text{FAC}_{i})$ is a \ac{FAC} input event at time $t^\text{FAC}_i$, and $w^\text{FAC}$ is a fixed weight.
The \ac{TRG} input events are similarly integrated, but the effective weight of the \ac{TRG} spikes is determined by the value of the FAC trace,  
\begin{equation}
    \frac{\text{d}I_{\text{TRG}}}{\text{d} t} = - \frac{I_{\text{TRG}}}{\tau_{\text{TRG}}} + I_{\text{FAC}}(t)\sum_{i} w^{\text{TRG}} \delta(t - t^\text{TRG}_{i})
\label{eqn:trg}
\end{equation}
where $\tau_\text{TRG}$ is the time constant of the decay, $\delta(t - t^\text{TRG}_{i})$ is a \ac{TRG} input spike at time $t^\text{TRG}_i$, $w^\text{TRG}$ a fixed weight, and $I_\text{FAC}(t)$ the value of the FAC trace at time $t$. 
The \ac{TRG} trace is then treated as an \ac{EPSC} into a \ac{LIF} neuron with time constant $\tau_m$ and firing threshold $u_\theta$, modeled as
%
\begin{equation}
    \frac{\text{d}u}{\text{d}t} = - \frac{u}{\tau_{m}} + I_{\text{TRG}}
\label{eqn:lif}
\end{equation}
where $u$ is the neuron's membrane potential.
When$u$ exceeds the firing threshold $u_\theta$, a spike is produced and $u$ is reset to \qty{0}{}.
This system of coupled ODEs was simulated in discrete time steps using the Euler method.
 
\subsection{Egomotion network}

A population of \ac{TDE} units was assembled across the visual field, with \ac{FAC} and \ac{TRG} connectivity configured to be sensitive to motion.
As illustrated in Fig.~\ref{fig:tde_pixels}, the \ac{FAC} and \ac{TRG} inputs can be mapped to the pixels from an event camera, in this example, the \ac{TDE} unit is sensitive to left-right motion and has a stride of 1: corresponding to the number of pixels separating the two inputs. 

The motion of static objects in a scene due to self-motion can be discerned by a network of \ac{TDE} units.
In this way, the mapping of a population of \acp{TDE} can be chosen arbitrarily or selected specifically to be receptive to certain degrees of motion across the entire visual field or segments of it.
In this work, the orientations of \ac{TDE} units were chosen to be either right-left or left-right sensitive. 
These orientations are therefore primarily receptive to the motion of an agent moving relative to its yaw axis, in other words, left and right turns.
The output spikes of each population of \acp{TDE}, corresponding to the two opposing orientations, were integrated to provide an estimate of the motion in their preferred direction.
This integration was modeled as a leaky integrator where events from left-right orientated \acp{TDE}, $\delta(t - t_{i}^{L\rightarrow R})$, increment the activity variable, $A$, and right-left events, $\delta(t - t_{i}^{R\rightarrow L})$, decrement it. 
The time constant of this process is given by $\tau_A$.
\begin{equation}
    \tau_A\frac{\mathrm{d}A}{\mathrm{d}t} = -A + \sum_{i}\delta(t - t_{i}^{L\rightarrow R})  - \sum_{i}\delta(t - t_{i}^{R\rightarrow L})
    \label{eq:leaky_int}
\end{equation}
%
This relative activity of the two populations was then normalised with a normalisation constant determined by the absolute maximum activity for the duration of the sample.
All experiments and simulations described in the preceding sections used the MVSEC dataset as input event data. 
Events of both polarities from a single event camera (the left camera in the stereoscopic setup) and its \ac{IMU} were used.

\subsubsection{On-chip}

A single \ac{TDE} circuit is realised on the \mbox{\textit{cognigr1}} \ac{ASIC}, therefore the network was run serially in time. 
The network has a feed-forward architecture therefore the response of each \ac{TDE} could be tested in isolation and the network could be executed sequentially.
Given this constraint and the large number of pixels within the visual field available to be sampled from, a spatially downsampled network configuration was deployed.
Fig.~\ref{fig:on-chip_setup} illustrates the setup for the egomotion network that was executed on-chip. 
Two sample areas were chosen to be \mbox{$20\times20$} in size, referred to as left and right boxes. 
Both boxes were positioned in the vertical center of the visual field but offset in the $x$-direction by 100 pixels ($\sim 30\%$ from the center).
Within each box, a population of 100 \ac{TDE} units of half left-right orientation and half right-left orientation was randomly distributed with a stride of 2 pixels, this distribution was identical for both boxes \mbox{(Fig.~\ref{fig:on-chip_setup}a,c)}. 
This configuration was chosen arbitrarily to process a reasonable subspace of the input events, excluding, for example, events generated by the car's bonnet.
Within each box, the differential activity (Eq.\ref{eq:leaky_int}) of the opposing \ac{TDE} orientations was computed and then summed for both boxes, resulting in a signal representative of the self-motion of the vehicle. 
For the on-chip experiments, the biasing of the \ac{TDE} circuitry was fixed, these biases are listed in Table~\ref{tab:circuit_parameters}. 
Additionally, the power consumption of the isolated \ac{TDE} and neuron circuit was measured for a random 40 runs using an \ac{SMU}.
The identical network configuration with the same input data was also simulated using the \textit{Brian2} simulator~\cite{stimberg_brian_2019}, the parameters of this simulation are provided in Table~\ref{tab:brian2_parameters}.

\subsubsection{Simulation}

To analyse the effect of scaling up the egomotion network—an approach unfeasible due to the spatial resolution of state-of-the-art event cameras and the limitation of a single \ac{TDE} circuit on the chip—it was simulated in discrete time on a \ac{GPU} using the \textit{JAX} library.
This approach enabled dense sampling of the entire visual field, as illustrated in \mbox{Fig.~\ref{fig:on-chip_setup}b}, and allowed for egomotion estimation to be obtained as if a large array of \ac{TDE} units were implemented on-chip.
This network utilised 178,880 units with an equal split of each \ac{TDE} orientation and a stride of 2 pixels. 
The \ac{TDE} model parameters used in these simulations are provided in Supplementary Table~\ref{tab:jax_parameters}.


\section{Acknowledgments}

This work has been supported by DFG projects MemTDE (441959088) and NMVAC (432009531), and EU H2020 projects NeuTouch (813713).
Funded by the Deutsche Forschungsgemeinschaft (DFG German Research Foundation) - Project number 432009531.
O.R. and M.M. were solely affiliated with BICS\textsuperscript{1} and CogniGron\textsuperscript{2} during the time of this work. 
The authors would like to acknowledge the financial support of the CogniGron research center and the Ubbo Emmius Funds (Univ. of Groningen). 
The authors would also like to acknowledge Philipp Klein and Ton Juny Pina for soldering the PCBs.

\section{Contributions}

Conceptualisation - H. G., E.C.; 
Methodology - H. G., M. C., M. M., O. R., E.C.; 
Software/Hardware - H. G., M. M., O. R.; 
Investigation - H. G., M. M., E.C.; 
Writing - H. G., M. C., M. M., O. R., E.C.; 
Visualisation - H. G., M. C.;  
Supervision - E. C.

\FloatBarrier
\balance

\section*{References}
\addcontentsline{toc}{section}{References}

\widowpenalty=10000
\clubpenalty=10000

\printbibliography[heading=none]

\end{refsection}

\begin{refsection}
    




\renewcommand\thefigure{S\arabic{figure}}
\renewcommand\thesection{S\arabic{section}}
\renewcommand{\theHfigure}{S\arabic{figure}}
\renewcommand\theHsection{S\arabic{section}}

\onecolumn


\begin{center}
\textsc{Supplementary Material}
\end{center}

\setcounter{figure}{0}
\setcounter{section}{0}
\setcounter{equation}{0}
\setcounter{page}{1}

\section{TDE synapse circuit analysis}
\label{sec:circuit_analysis}

 The following section refers to \mbox{Fig.~\ref{fig:cognigr1_shem}b} and the currents, capacitances and voltages shown on the schematic. 
 This approach to characterise the behaviour of the \ac{TDE} circuit is derived from similar analysis of \ac{DPI} circuits found in literature \cite{bartolozzi2007}.
 In order to derive the dynamics of $V^{\text{FAC}}$, an expression for the current flow, $I_1$, as a result of a digital pulse arriving at the gate transistor $\text{M}_1$ is required.
 The following assumes that the subthreshold slopes of the \ac{MOS} transistors is the same for \ac{PMOS} and \ac{NMOS} transistors, $k_\mathrm{n}\approx k_\mathrm{p} \approx k$. 
 The subthreshold currents through transistors $\text{M}_2$ and $\text{M}_7$ are given by the following equations. 

 \begin{equation}
    I_\mathrm{gain}^\mathrm{FAC} = I_0 e^{-\frac{\kappa}{U_T} V^\mathrm{FAC}_{\mathrm{gain}}}
\end{equation}

\begin{equation}
    I^{\mathrm{TRG}}_{\mathrm{gain}} = I_0 e^{-\frac{\kappa}{U_T} V^{\mathrm{FAC}}}
    \label{eq:I_trg_gain}
\end{equation}

The current $I_1$ given by the behaviour of the \ac{DPI} and the biasing of the circuit's transistors can be written and refactored as follows. 

\begin{equation}
    I_1 = I_{\mathrm{w}}^\mathrm{FAC} \frac{e^{\frac{\kappa V^{\mathrm{FAC}}}{U_T}}}{e^{\frac{\kappa V^{\mathrm{FAC}}}{U_T}} + e^{\frac{\kappa V^\mathrm{FAC}_{\mathrm{gain}}}{U_T}}}
\end{equation}

\begin{equation}
    I_1 = \frac{I_{\mathrm{w}}^\mathrm{FAC}}{1 + \frac{I_\mathrm{gain}^\mathrm{TRG}}{I_\mathrm{gain}^\mathrm{FAC}}}
\end{equation}

The voltage $V^\mathrm{FAC}$ is determined by the movement of charge on capacitor $C^{\mathrm{FAC}}$. 
These dynamics are governed by the following \ac{DE}, where $I^\mathrm{FAC}_\tau$ is a constant leak controlled by the biasing of transistor $\text{M}_4$, $V^\mathrm{FAC}_{\tau}$. 

\begin{equation}
    C^{\mathrm{FAC}} \frac{\mathrm{d}}{\mathrm{d}t} V^{\mathrm{FAC}} = (I_1 - I^\mathrm{FAC}_\tau)
\end{equation}

Additionally, the current flowing through \ac{PMOS} transistor $\text{M}_4$, operating in the threshold regime is given by the following equation.

\begin{equation}
    I^{\mathrm{FAC}}_{\tau} = I_0 e^{-\frac{\kappa}{U_T} (V^\mathrm{FAC}_{\tau} - V_{\mathrm{dd}})}
\end{equation}

Taking the time derivative of equation~\ref{eq:I_trg_gain} and equating with equation~\ref{eq:dt_v_fac}.

\begin{equation}
     \frac{\mathrm{d}}{\mathrm{d}t} V^{\mathrm{FAC}} = - \frac{1}{C^{\mathrm{FAC}}}(I_1 - I^\mathrm{FAC}_\tau)
\end{equation}

\begin{equation}
     \frac{\mathrm{d}}{\mathrm{d}t} V^{\mathrm{FAC}} = - \frac{U_T}{\kappa}\frac{1}{I^\mathrm{TRG}_{\mathrm{gain}}} \frac{\mathrm{d}I^\mathrm{TRG}_{\mathrm{gain}}}{\mathrm{d}t}
     \label{eq:dt_v_fac}
\end{equation}

\begin{equation}
    \frac{\mathrm{d}I^\mathrm{TRG}_\mathrm{gain}}{\mathrm{d}t} = \frac{I^\mathrm{TRG}_{\mathrm{gain}} \kappa}{U_T C^{\mathrm{FAC}}}(-I_1 - I^\mathrm{FAC}_\tau )
\end{equation}

Substituting $I_1$

\begin{equation}
    \frac{\mathrm{d}I^\mathrm{TRG}_\mathrm{gain}}{\mathrm{d}t} = \frac{I^\mathrm{TRG}_{\mathrm{gain}} \kappa}{U_T C^{\mathrm{FAC}}} \Bigg ( \frac{I_{\mathrm{w}}^\mathrm{FAC}}{1 + \frac{I^\mathrm{TRG}_{\mathrm{gain}}}{I_\mathrm{gain}^\mathrm{FAC}}} - I^\mathrm{FAC}_\tau \Bigg )
\end{equation}

Collecting terms and defining $\tau_{\mathrm{FAC}}$

\begin{equation}
    \tau_{\mathrm{FAC}} \frac{\mathrm{d}I^\mathrm{TRG}_{\mathrm{gain}}}{\mathrm{d}t} =  I^\mathrm{TRG}_{\mathrm{gain}} \Bigg (\frac{1}{I^\mathrm{FAC}_\tau}\frac{I_{\mathrm{w}}^\mathrm{FAC}}{1 + \frac{I^\mathrm{TRG}_{\mathrm{gain}}}{I_\mathrm{gain}^\mathrm{FAC}}} - 1 \Bigg )
    \label{eq:one}
\end{equation}

\begin{equation}
    \tau_\mathrm{FAC} = \frac{U_T C^\mathrm{FAC}}{\kappa I^\mathrm{FAC}_\tau}
\end{equation}

If $I_{\mathrm{w}}^\mathrm{FAC} >> I_{\mathrm{FAC}_\tau}$, then $I_\mathrm{gain}^\mathrm{FAC}$ will eventually rise to such that $I_\mathrm{gain}^\mathrm{TRG} >> I_\mathrm{gain}^\mathrm{FAC}$, when the circuit is stimulated with a step signal. If $\frac{I_\mathrm{gain}^\mathrm{TRG}}{I_\mathrm{gain}^\mathrm{FAC}} >> 1$ the $I_\mathrm{gain}^\mathrm{TRG}$ dependence in \ref{eq:one} cancels out. 

\begin{equation}
    \tau_{\mathrm{FAC}} \frac{\mathrm{d}I^\mathrm{TRG}_{\mathrm{gain}}}{\mathrm{d}t} + I^\mathrm{TRG}_{\mathrm{gain}} =  \frac{I_{\mathrm{w}}^\mathrm{FAC}}{I^\mathrm{FAC}_\tau}I_\mathrm{gain}^\mathrm{FAC}
\end{equation}

The response of $I^\mathrm{TRG}_{\mathrm{gain}}$ arrive at $\mathrm{FAC}$ at $t^{-}$ and ending at $t^+$ is

\begin{equation}
    \begin{aligned}
        I^\mathrm{TRG}_{\mathrm{gain}}(t) = 
        \begin{cases}
        \frac{{I_{\mathrm{w}}^\mathrm{FAC}}I^\mathrm{FAC}_{\mathrm{gain}}}{I^\mathrm{FAC}_\tau} \bigg (1- e^{-\frac{(t-t^-)}{\tau_{\mathrm{FAC}}}}\bigg ) + I^\mathrm{TRG-}_{\mathrm{gain}} e^{-\frac{(t-t^-)}{\tau_{\mathrm{FAC}}}} \ \mathrm{(discharge \ phase)} \\ \noalign{\vskip9pt} I^\mathrm{TRG-}_{\mathrm{gain}} + \big (I^\mathrm{TRG+}_{\mathrm{gain}} - I^\mathrm{TRG-}_{\mathrm{gain}} \big ) e^{-\frac{(t-t^+)}{\tau_{\mathrm{FAC}}}} \ \mathrm{(charge \ phase)}
        \end{cases}
    \end{aligned}
\end{equation}

\begin{equation}
    \tau_{\mathrm{TRG}} \frac{\mathrm{d}I^{\mathrm{TDE}}}{\mathrm{d}t} =  I^{\mathrm{TDE}} \bigg ( 1 - \frac{1}{I^\mathrm{TRG}_\tau}\frac{I^\mathrm{TRG}_\mathrm{w}}{1 + \frac{I^{\mathrm{TDE}}}{I^\mathrm{TRG}_{\mathrm{gain}}(t)}} \bigg )
\end{equation}

\begin{equation}
    \tau_{\mathrm{TRG}} \frac{\mathrm{d}I^{\mathrm{TDE}}}{\mathrm{d}t} - I^{\mathrm{TDE}} = - \frac{I^\mathrm{TRG}_\mathrm{w}}{I^\mathrm{TRG}_\tau}I^\mathrm{TRG}_{\mathrm{gain}}(t)
\end{equation}

\begin{equation}
    \begin{aligned}
        \tau_{\mathrm{TRG}} \frac{\mathrm{d}I^{\mathrm{TDE}}}{\mathrm{d}t} + I^{\mathrm{TDE}} = 
        \begin{cases}
         \frac{I^\mathrm{TRG}_\mathrm{w}}{I^\mathrm{TRG}_\tau}I^\mathrm{TRG-}_{\mathrm{gain}}  \ \mathrm{for} \  t < t^-\\ \noalign{\vskip9pt}
         \frac{I^\mathrm{TRG}_\mathrm{w}}{I^\mathrm{TRG}_\tau} \bigg [\frac{I_{\mathrm{w}}^\mathrm{FAC}I_\mathrm{gain}^\mathrm{FAC}}{I^\mathrm{FAC}_\tau} \bigg (1- e^{-\frac{(t-t^-)}{\tau_{\mathrm{FAC}}}} \bigg ) + I^\mathrm{TRG-}_{\mathrm{gain}} e^{-\frac{(t-t^-)}{\tau_{\mathrm{FAC}}}} \bigg ] \ \mathrm{for} \  t^- \leq t < t^+ \\ \noalign{\vskip9pt} \frac{I^\mathrm{TRG}_\mathrm{w}}{I^\mathrm{TRG}_\tau} \bigg [ I^\mathrm{TRG-}_{\mathrm{gain}} + \big (I^\mathrm{TRG+}_{\mathrm{gain}} - I^\mathrm{TRG-}_{\mathrm{gain}} \big ) e^{-\frac{(t-t^+)}{\tau_{\mathrm{FAC}}}} \bigg ] \ \mathrm{for} \ t \geq t^+
        \end{cases}
    \end{aligned}
\end{equation}

We analyse the case for $t \geq t^+$, translating to a trigger pulse arriving after the end of a facilitator pulse. 

\begin{equation}
    \tau_{\mathrm{TRG}} \frac{\mathrm{d}I^{\mathrm{TDE}}}{\mathrm{d}t} + I^{\mathrm{TDE}} =  \frac{I^\mathrm{TRG}_\mathrm{w}}{I^\mathrm{TRG}_\tau} \bigg [ I^\mathrm{TRG-}_{\mathrm{gain}} + \big (I^\mathrm{TRG+}_{\mathrm{gain}} - I^\mathrm{TRG-}_{\mathrm{gain}} \big ) e^{-\frac{t}{\tau_{\mathrm{FAC}}}} \bigg ]
\end{equation}

\begin{equation}
    \begin{aligned}
        I^{\mathrm{TDE}}(t) = 
        \begin{cases}

        \bigg [1- e^{-\frac{t - t^-_{\mathrm{TDE}}}{\tau_\mathrm{TRG}}}\bigg] \frac{I^\mathrm{TRG}_\mathrm{w}I^\mathrm{TRG-}_{\mathrm{gain}}}{I^\mathrm{TRG}_\tau} + I^\mathrm{TDE-}e^{-\frac{t - t^-_{\mathrm{TDE}}}{\tau_\mathrm{TRG}}}

        \\ \noalign{\vskip12pt}

        + \frac{\tau_{\mathrm{FAC}}}{\tau_{\mathrm{FAC}} - \tau_{\mathrm{TRG}}} \frac{I^\mathrm{TRG}_\mathrm{w}}{I^\mathrm{TRG}_\tau} \big (I^\mathrm{TRG+}_{\mathrm{gain}} - I^\mathrm{TRG-}_{\mathrm{gain}} \big ) \bigg [ e^{-\frac{t^-_{\mathrm{TDE}} - t}{\tau_\mathrm{FAC}}} - e^{-\frac{t - t^-_{\mathrm{TDE}}}{\tau_\mathrm{TRG}}}\bigg]  e^{\frac{-t^-_{\mathrm{TDE}}}{\tau_\mathrm{FAC}}} \ \mathrm{(discharge \ phase)}
         
         \\ \noalign{\vskip12pt} I^{\mathrm{TDE+}} e^{-\frac{(t - t^+_\mathrm{TDE})}{\tau_{\mathrm{TRG}}}}

         \ \mathrm{(charge \ phase)}
        \end{cases}
    \end{aligned}
\end{equation}

Consider discharge phase at $t = t^-_\mathrm{TDE} + \delta t = t^+_\mathrm{TDE}$  where $\delta t << t^-_\mathrm{TDE}$

\begin{equation}
    \begin{split}
        I^{\mathrm{TDE}}(t^-_\mathrm{TDE} + \delta t) \approx & -\frac{\delta t}{\tau_\mathrm{TRG}} \frac{I^\mathrm{TRG}_\mathrm{w}I^\mathrm{TRG-}_{\mathrm{gain}}}{I^\mathrm{TRG}_\tau} + I^\mathrm{TDE-}(1 -\frac{\delta t}{\tau_\mathrm{TRG}}) \\ & + \frac{\tau_{\mathrm{FAC}}}{\tau_{\mathrm{FAC}} - \tau_{\mathrm{TRG}}} \frac{I^\mathrm{TRG}_\mathrm{w}}{I^\mathrm{TRG}_\tau} \big (I^\mathrm{TRG-}_{\mathrm{gain}} - I^\mathrm{TRG+}_{\mathrm{gain}} \big ) \bigg [\frac{\delta t}{\tau_\mathrm{FAC}} - \frac{\delta t}{\tau_\mathrm{TRG}} \bigg ]  e^{\frac{-t^-_{\mathrm{TDE}}}{\tau_\mathrm{FAC}}}        
    \end{split}
\end{equation}

If we assume $I^\mathrm{TDE-} \approx 0$ the result is a dependence on the time of arrival of the facilitator pulse $t^-_{\mathrm{TDE}}$ on the maximum current drawn from $\mathrm{M}_{10}$.

\begin{equation}
    I^\mathrm{TDE+} \propto  e^{\frac{-t^-_{\mathrm{TDE}}}{\tau_\mathrm{FAC}}}
\end{equation}

\section{ARRE}
\label{sec:arre}
The \ac{ARRE} is defined as follows: $P$ is a vector of the predicted Euler angles and $G$ the ground truth, at each measurement time step. 
$\mathrm{logm}(\cdot)$ is matrix logarithm and $||\cdot||_2$ is the $L2$ norm.

\begin{equation}
    \text{ARRE}(P, G) = \frac{1}{n}\sum^n_{i=1} || \mathrm{logm}(P_i^TG_i)||_2
\end{equation}




\section{Silicon Neuron}
\label{sec:neuron}


\begin{figure}[H]
    \centering
    \includegraphics[width=0.6\linewidth]{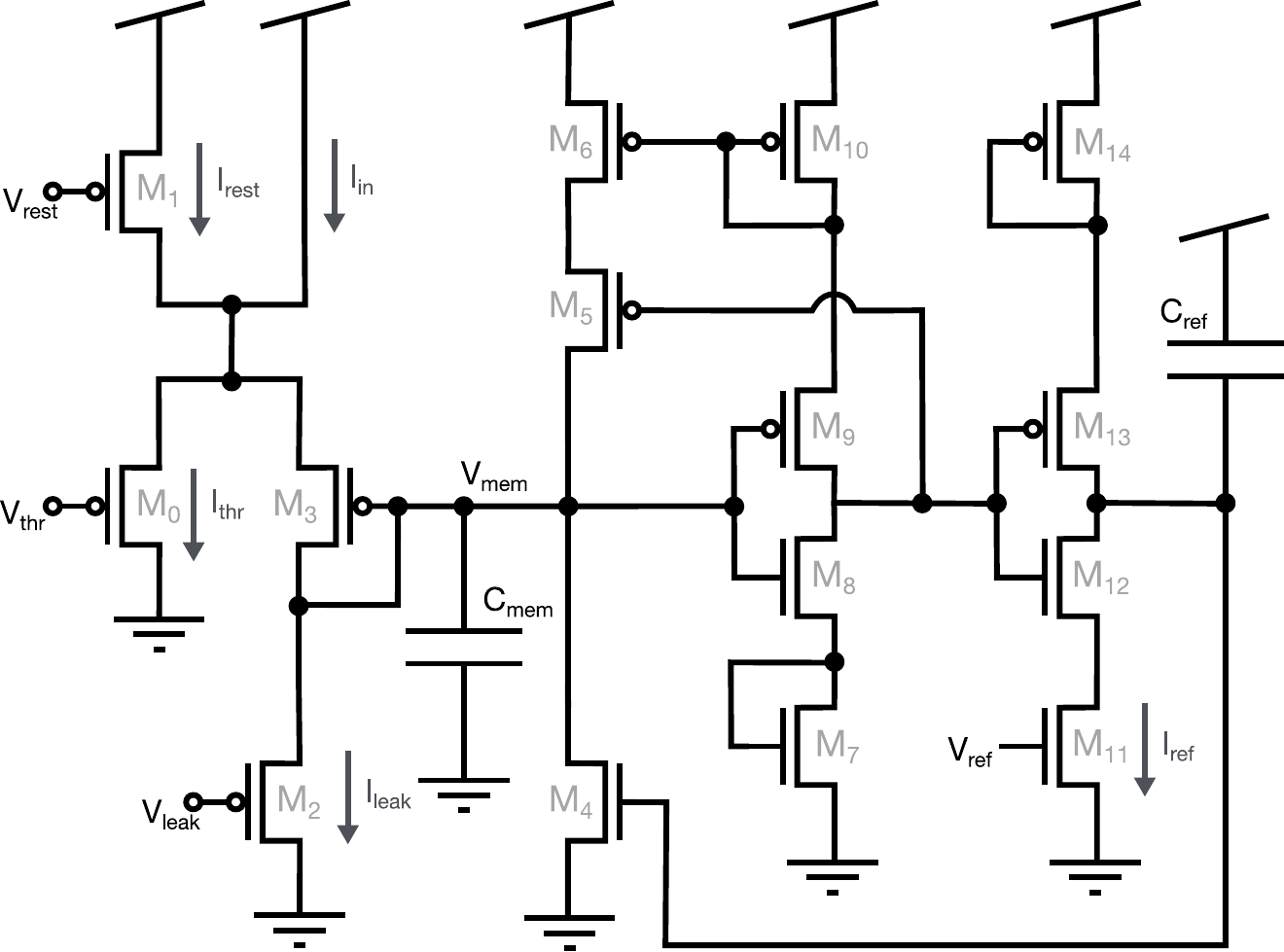}
    \vspace{10pt}
    \caption{\textbf{The silicon neuron circuit used in the \ac{TDE} implementation on the \textit{cognigr1} \ac{ASIC} \cite{richter_subthreshold_2023, cotteret_robust_2023, mastella_synaptic_2023}.}
    Within the neuron, the input current results from the combination of $I_\text{in}$ and $I_\text{rest}$. 
    $I_\text{rest}$ is a current employed to establish the resting voltage level across $C_\text{mem}$, representing the membrane potential $V_\text{mem}$. 
    The inverter, $\text{M}_8-\text{M}_9$, switches when the membrane voltage reaches a its switching threshold. 
    Upon switching, the current passing through it, $I_\text{mem}$, increases. 
    This consequently triggers a positive feedback loop through the current mirror $\text{M}_6-\text{M}_{10}$, causing the amplification current $I_\text{mem}$, to further increase $V_\text{mem}$. 
    As the output of the initial inverter transitions to a low state, the second inverter, $\text{M}_{13}-\text{M}_{12}$, initiates the charging of capacitor $C_\text{ref}$. 
    The voltage on $C_\text{ref}$ triggers $\text{M}_4$ to become conductive, leading to the discharge of $C_\text{mem}$ with a current $I_\text{ref}$, ultimately resetting the neuron. 
    However, due to the time required for the capacitor to charge, this negative feedback mechanism operates more slowly than the positive feedback, resulting in the generation of a spike. 
    Following the discharge of $V_\text{mem}$ until the output of the first inverter returns to a high state, $C_\text{ref}$ undergoes a gradual discharge through $\text{M}_4$, with a current determined by $V_\text{ref}$. 
    This establishes a designated refractory period, during which $C_\text{mem}$ experiences continuous discharge through $\text{M}_4$ meaning that the neuron circuit cannot generate a spike.}
    \vspace{-1em}
    \label{fig:neuron}
\end{figure}

\section{Parameters}
\label{sec:biases}

\begin{figure}[H]
\begin{subfigure}[b]{0.2\textwidth}
\renewcommand{\arraystretch}{1.2}
\centering
\begin{tabular}{ll}
\toprule
Circuit bias & Value \\ \hline
$I^\text{TRG}_\tau$ & $2 \ \text{pA}$\\
$I^\text{TRG}_{\text{w}}$ & $4 \ \text{nA}$ \\
$I^\text{FAC}_\tau$ & $2 \ \text{pA}$ \\
$I^\text{FAC}_{\text{w}}$ & $4 \ \text{nA}$ \\
$I^\text{FAC}_{\text{gain}}$ & $10 \ \text{pA}$ \\
$I_{\text{ref}}$ & $2 \ \text{pA}$ \\
$I_{\text{leak}}$& $2 \ \text{pA}$ \\
$I_{\text{rest}}$ & $2 \ \text{pA}$ \\
$I_{\text{thr}}$ & $2 \ \text{pA}$ \\
\bottomrule
\end{tabular}
\caption{\ac{TDE} circuit}
\label{tab:circuit_parameters}
\end{subfigure}
\hfill
\begin{subfigure}[b]{0.35\textwidth}
\centering
\renewcommand{\arraystretch}{1.2}
\centering
\begin{tabular}{ll}
\toprule
Simulation parameter & Value \\ \hline
$u_{\theta}$ & $50$ \\
$\tau_{\text{FAC}}$ & $20$ ms\\
$\tau_{\text{TRG}}$ & $20$ ms\\
$\tau_{m}$ & $20$ ms\\
$w^{\text{FAC}}$ & $1$ \\
$w^{\text{TRG}}$ & $1$ \\
$\tau_{A}$ & $0.75$ s\\
$t_{\text{ref}}$ & $0.1$ ms\\
\bottomrule
\end{tabular}
\vspace{15pt}
\caption{Brian2}
\label{tab:brian2_parameters}
\end{subfigure}
\begin{subfigure}[b]{0.35\textwidth}
\renewcommand{\arraystretch}{1.2}
\centering
\begin{tabular}{ll}
\toprule
Simulation parameter & Value \\ \hline
$u_{\theta}$ & $50$\\
$\tau_{\text{FAC}}$ & $20$ ms\\
$\tau_{\text{TRG}}$ & $20$ ms\\
$\tau_{m}$ & $20$ ms\\
$w^{\text{FAC}}$ & $1$ \\
$w^{\text{TRG}}$ & $1$ \\
$\tau_{A}$ & $10$ ms \\
\bottomrule
\end{tabular}
\vspace{32pt}
\caption{JAX}
\label{tab:jax_parameters}
\end{subfigure}
\caption{\textbf{a)} Circuit biases used for all on-chip \ac{TDE} measurements and experiments. 
\textbf{b)} Parameters used in \textit{Brian2} simulations. \textbf{c)} Parameters used in \textit{JAX} simulations.}
\end{figure}

\printbibliography[title={Supplementary references}]

@article{stimberg_brian_2019,
    title = {Brian 2, an intuitive and efficient neural simulator},
    volume = {8},
    issn = {2050-084X},
    doi = {10.7554/eLife.47314},
    journal = {eLife},
    author = {Stimberg, Marcel and Brette, Romain and Goodman, Dan FM},
    editor = {Skinner, Frances K},
    month = aug,
    year = {2019},
    pages = {e47314}
}

@article{lichtsteiner2008_fix,
	title = {A 128 × 128 120 {dB} 15 $\mu$s latency asynchronous temporal contrast vision sensor},
	volume = {43},
	issn = {00189200},
	doi = {10.1109/JSSC.2007.914337},
	number = {2},
	urldate = {2023-07-19},
	journal = {IEEE Journal of Solid-State Circuits},
	author = {Lichtsteiner, Patrick and Posch, Christoph and Delbruck, Tobi},
	month = feb,
	year = {2008},
	pages = {566--576},
}

@article{schoepe2023,
	title = {Closed-loop sound source localization in neuromorphic systems},
	volume = {3},
	issn = {26344386},
	doi = {10.1088/2634-4386/ACDABA},
	number = {2},
	urldate = {2024-01-15},
	journal = {Neuromorphic Computing and Engineering},
	author = {Schoepe, Thorben and Gutierrez-Galan, Daniel and Dominguez-Morales, Juan P. and Greatorex, Hugh and Jimenez-Fernandez, Angel and Linares-Barranco, Alejandro and Chicca, Elisabetta},
	month = jun,
	year = {2023},
	note = {Publisher: Institute of Physics},
}

@inproceedings{paredes-valles_taming_2023,
	title = {Taming {Contrast} {Maximization} for {Learning} {Sequential}, {Low}-latency, {Event}-based {Optical} {Flow}},
	url = {https://openaccess.thecvf.com/content/ICCV2023/html/Paredes-Valles_Taming_Contrast_Maximization_for_Learning_Sequential_Low-latency_Event-based_Optical_Flow_ICCV_2023_paper.html},
	language = {en},
	urldate = {2024-12-05},
	booktitle = {Proceedings of the {IEEE}/{CVF} {International} {Conference} on {Computer} {Vision}},
	author = {Paredes-Vallés, Federico and Scheper, Kirk Y. W. and De Wagter, Christophe and de Croon, Guido C. H. E.},
	year = {2023},
	pages = {9695--9705},
}

@inproceedings{amir_low_2017,
	title = {A {Low} {Power}, {Fully} {Event}-{Based} {Gesture} {Recognition} {System}},
	url = {https://openaccess.thecvf.com/content_cvpr_2017/html/Amir_A_Low_Power_CVPR_2017_paper.html},
	urldate = {2024-12-10},
	booktitle = {Proceedings of the {IEEE} {Conference} on {Computer} {Vision} and {Pattern} {Recognition}},
	author = {Amir, Arnon and Taba, Brian and Berg, David and Melano, Timothy and McKinstry, Jeffrey and Di Nolfo, Carmelo and Nayak, Tapan and Andreopoulos, Alexander and Garreau, Guillaume and Mendoza, Marcela and Kusnitz, Jeff and Debole, Michael and Esser, Steve and Delbruck, Tobi and Flickner, Myron and Modha, Dharmendra},
	year = {2017},
	pages = {7243--7252},
}

@misc{greatorex_scalable_2025,
	title = {A scalable event-driven spatiotemporal feature extraction circuit},
	url = {http://arxiv.org/abs/2501.10155},
	doi = {10.48550/arXiv.2501.10155},
	urldate = {2025-01-20},
	publisher = {arXiv},
	author = {Greatorex, Hugh and Mastella, Michele and Richter, Ole and Cotteret, Madison and Girão, Willian Soares and Janotte, Ella and Chicca, Elisabetta},
	month = jan,
	year = {2025},
	note = {arXiv:2501.10155 [eess]},
}

@article{yang_evgnn_2024,
	title = {{EvGNN}: {An} {Event}-driven {Graph} {Neural} {Network} {Accelerator} for {Edge} {Vision}},
	issn = {2996-6647},
	shorttitle = {{EvGNN}},
	url = {https://ieeexplore.ieee.org/document/10812004},
	doi = {10.1109/TCASAI.2024.3520905},
	urldate = {2025-01-16},
	journal = {IEEE Transactions on Circuits and Systems for Artificial Intelligence},
	author = {Yang, Yufeng and Kneip, Adrian and Frenkel, Charlotte},
	year = {2024},
	note = {Conference Name: IEEE Transactions on Circuits and Systems for Artificial Intelligence},
	pages = {1--14},
}

@article{borenstein_mobile_1997,
	title = {Mobile robot positioning: {Sensors} and techniques},
	volume = {14},
	copyright = {Copyright © 1997 John Wiley \& Sons, Inc.},
	issn = {1097-4563},
	shorttitle = {Mobile robot positioning},
	url = {https://onlinelibrary.wiley.com/doi/abs/10.1002/%28SICI%291097-4563%28199704%2914%3A4%3C231%3A%3AAID-ROB2%3E3.0.CO%3B2-R},
	doi = {10.1002/(SICI)1097-4563(199704)14:4<231::AID-ROB2>3.0.CO;2-R},
	language = {en},
	number = {4},
	urldate = {2025-01-14},
	journal = {Journal of Robotic Systems},
	author = {Borenstein, J. and Everett, H. R. and Feng, L. and Wehe, D.},
	year = {1997},
	note = {\_eprint: https://onlinelibrary.wiley.com/doi/pdf/10.1002/\%28SICI\%291097-4563\%28199704\%2914\%3A4\%3C231\%3A\%3AAID-ROB2\%3E3.0.CO\%3B2-R},
	pages = {231--249},
}

@inproceedings{mitrokhin_ev-imo_2019,
	title = {{EV}-{IMO}: {Motion} {Segmentation} {Dataset} and {Learning} {Pipeline} for {Event} {Cameras}},
	shorttitle = {{EV}-{IMO}},
	url = {https://ieeexplore.ieee.org/abstract/document/8968520},
	doi = {10.1109/IROS40897.2019.8968520},
	urldate = {2025-01-13},
	booktitle = {2019 {IEEE}/{RSJ} {International} {Conference} on {Intelligent} {Robots} and {Systems} ({IROS})},
	author = {Mitrokhin, Anton and Ye, Chengxi and Fermüller, Cornelia and Aloimonos, Yiannis and Delbruck, Tobi},
	month = nov,
	year = {2019},
	note = {ISSN: 2153-0866},
	pages = {6105--6112},
}

@article{benosman_event-based_2014,
	title = {Event-{Based} {Visual} {Flow}},
	volume = {25},
	issn = {2162-2388},
	url = {https://ieeexplore.ieee.org/abstract/document/6589170},
	doi = {10.1109/TNNLS.2013.2273537},
	number = {2},
	urldate = {2025-01-13},
	journal = {IEEE Transactions on Neural Networks and Learning Systems},
	author = {Benosman, Ryad and Clercq, Charles and Lagorce, Xavier and Ieng, Sio-Hoi and Bartolozzi, Chiara},
	month = feb,
	year = {2014},
	note = {Conference Name: IEEE Transactions on Neural Networks and Learning Systems},
	pages = {407--417},
}

@article{haessig_spiking_2018,
	title = {Spiking {Optical} {Flow} for {Event}-{Based} {Sensors} {Using} {IBM}'s {TrueNorth} {Neurosynaptic} {System}},
	volume = {12},
	issn = {1940-9990},
	url = {https://ieeexplore.ieee.org/abstract/document/8388703},
	doi = {10.1109/TBCAS.2018.2834558},
	number = {4},
	urldate = {2025-01-13},
	journal = {IEEE Transactions on Biomedical Circuits and Systems},
	author = {Haessig, Germain and Cassidy, Andrew and Alvarez, Rodrigo and Benosman, Ryad and Orchard, Garrick},
	month = aug,
	year = {2018},
	note = {Conference Name: IEEE Transactions on Biomedical Circuits and Systems},
	pages = {860--870},
}

@article{palossi_64-mw_2019,
	title = {A 64-{mW} {DNN}-{Based} {Visual} {Navigation} {Engine} for {Autonomous} {Nano}-{Drones}},
	volume = {6},
	issn = {2327-4662},
	url = {https://ieeexplore.ieee.org/abstract/document/8715489?casa_token=AiPxYHM6Du0AAAAA:qOn6XJebEbzqwhgTAtexxHJDkknsS53E6TKMp3SMGxjs-cOgJ3wil4GN4qr7Irjcm6uQn2qHJs0},
	doi = {10.1109/JIOT.2019.2917066},
	number = {5},
	urldate = {2024-12-18},
	journal = {IEEE Internet of Things Journal},
	author = {Palossi, Daniele and Loquercio, Antonio and Conti, Francesco and Flamand, Eric and Scaramuzza, Davide and Benini, Luca},
	month = oct,
	year = {2019},
	note = {Conference Name: IEEE Internet of Things Journal},
	pages = {8357--8371},
}

@inproceedings{nister_visual_2004,
	title = {Visual odometry},
	volume = {1},
	url = {https://ieeexplore.ieee.org/abstract/document/1315094?casa_token=BZrmc_YXqYYAAAAA:Ux1yLhqW_d61O4Z61TGN5Rp2o5K5BuT989v9gD7Y8bx0na7aNO_SONBRO-mp0LbiY3FTO68SkKM},
	doi = {10.1109/CVPR.2004.1315094},
	urldate = {2024-12-18},
	booktitle = {Proceedings of the 2004 {IEEE} {Computer} {Society} {Conference} on {Computer} {Vision} and {Pattern} {Recognition}, 2004. {CVPR} 2004.},
	author = {Nister, D. and Naroditsky, O. and Bergen, J.},
	month = jun,
	year = {2004},
	note = {ISSN: 1063-6919},
	pages = {I--I},
}

@inproceedings{chiavazza_low-latency_2023,
	title = {Low-{Latency} {Monocular} {Depth} {Estimation} {Using} {Event} {Timing} on {Neuromorphic} {Hardware}},
	booktitle = {Proceedings of the {IEEE}/{CVF} {Conference} on {Computer} {Vision} and {Pattern} {Recognition} ({CVPR}) {Workshops}},
	author = {Chiavazza, Stefano and Meyer, Svea Marie and Sandamirskaya, Yulia},
	month = jun,
	year = {2023},
	pages = {4071--4080},
}

@article{dangelo_event-based_2020,
	title = {Event-{Based} {Eccentric} {Motion} {Detection} {Exploiting} {Time} {Difference} {Encoding}},
	volume = {14},
	issn = {1662-453X},
	url = {https://www.frontiersin.org/journals/neuroscience/articles/10.3389/fnins.2020.00451},
	doi = {10.3389/fnins.2020.00451},
	journal = {Frontiers in Neuroscience},
	author = {D'Angelo, Giulia and Janotte, Ella and Schoepe, Thorben and O'Keeffe, James and Milde, Moritz B. and Chicca, Elisabetta and Bartolozzi, Chiara},
	year = {2020},
}

@inproceedings{richter_subthreshold_2023,
	address = {New York, NY, USA},
	series = {{ICONS} '23},
	title = {A {Subthreshold} {Second}-{Order} {Integration} {Circuit} for {Versatile} {Synaptic} {Alpha} {Kernel} and {Trace} {Generation}},
	isbn = {9798400701757},
	url = {https://doi.org/10.1145/3589737.3606008},
	doi = {10.1145/3589737.3606008},
	booktitle = {Proceedings of the 2023 {International} {Conference} on {Neuromorphic} {Systems}},
	publisher = {Association for Computing Machinery},
	author = {Richter, Ole and Greatorex, Hugh and Hucko, Benjamin and Cotteret, Madison and Soares Girao, Willian and Janotte, Ella and Mastella, Michele and Chicca, Elisabetta},
	year = {2023},
	note = {event-place: Santa Fe, NM, USA},
}

@inproceedings{mastella_synaptic_2023,
	address = {New York, NY, USA},
	series = {{ICONS} '23},
	title = {Synaptic {Normalisation} for {On}-{Chip} {Learning} in {Analog} {CMOS} {Spiking} {Neural} {Networks}},
	isbn = {9798400701757},
	url = {https://doi.org/10.1145/3589737.3606007},
	doi = {10.1145/3589737.3606007},
	booktitle = {Proceedings of the 2023 {International} {Conference} on {Neuromorphic} {Systems}},
	publisher = {Association for Computing Machinery},
	author = {Mastella, Michele and Greatorex, Hugh and Cotteret, Madison and Janotte, Ella and Soares Girao, Willian and Richter, Ole and Chicca, Elisabetta},
	year = {2023},
	note = {event-place: Santa Fe, NM, USA},
}

@inproceedings{cotteret_robust_2023,
	title = {Robust {Spiking} {Attractor} {Networks} with a {Hard} {Winner}-{Take}-{All} {Neuron} {Circuit}},
	doi = {10.1109/ISCAS46773.2023.10181513},
	booktitle = {2023 {IEEE} {International} {Symposium} on {Circuits} and {Systems} ({ISCAS})},
	author = {Cotteret, Madison and Richter, Ole and Mastella, Michele and Greatorex, Hugh and Janotte, Ella and Girão, Willian Soares and Ziegler, Martin and Chicca, Elisabetta},
	year = {2023},
	pages = {1--5},
}

@article{reichardt_autokorrelations-auswertung_1957,
	title = {Autokorrelations-{Auswertung} als {Funktionsprinzip} des {Zentralnervensystems}: (bei der optischen {Bewegungswahrnehmung} eines {Insektes})},
	volume = {12},
	copyright = {De Gruyter expressly reserves the right to use all content for commercial text and data mining within the meaning of Section 44b of the German Copyright Act.},
	issn = {1865-7117},
	shorttitle = {Autokorrelations-{Auswertung} als {Funktionsprinzip} des {Zentralnervensystems}},
	url = {https://www.degruyter.com/document/doi/10.1515/znb-1957-0707/html},
	doi = {10.1515/znb-1957-0707},
	language = {en},
	number = {7},
	urldate = {2024-12-13},
	journal = {Zeitschrift für Naturforschung B},
	author = {Reichardt, Werner},
	month = jul,
	year = {1957},
	note = {Publisher: De Gruyter},
	pages = {448--457},
}

@article{chen_event-based_2020,
	title = {Event-{Based} {Neuromorphic} {Vision} for {Autonomous} {Driving}: {A} {Paradigm} {Shift} for {Bio}-{Inspired} {Visual} {Sensing} and {Perception}},
	volume = {37},
	issn = {1558-0792},
	shorttitle = {Event-{Based} {Neuromorphic} {Vision} for {Autonomous} {Driving}},
	url = {https://ieeexplore.ieee.org/abstract/document/9129849?casa_token=neEoytuYC2sAAAAA:Pj62_e4n-U-mRMFaHCLtabaKKac3zGrU9Yk55aZ1F98IYxe2E7rFenrYSqrB-8FYnNEgwkiBes0},
	doi = {10.1109/MSP.2020.2985815},
	number = {4},
	urldate = {2024-12-10},
	journal = {IEEE Signal Processing Magazine},
	author = {Chen, Guang and Cao, Hu and Conradt, Jorg and Tang, Huajin and Rohrbein, Florian and Knoll, Alois},
	month = jul,
	year = {2020},
	note = {Conference Name: IEEE Signal Processing Magazine},
	pages = {34--49},
}

@article{shiba_secrets_2024,
	title = {Secrets of {Event}-{Based} {Optical} {Flow}, {Depth} and {Ego}-{Motion} {Estimation} by {Contrast} {Maximization}},
	volume = {46},
	issn = {1939-3539},
	url = {https://ieeexplore.ieee.org/abstract/document/10517639},
	doi = {10.1109/TPAMI.2024.3396116},
	number = {12},
	urldate = {2024-12-05},
	journal = {IEEE Transactions on Pattern Analysis and Machine Intelligence},
	author = {Shiba, Shintaro and Klose, Yannick and Aoki, Yoshimitsu and Gallego, Guillermo},
	month = dec,
	year = {2024},
	note = {Conference Name: IEEE Transactions on Pattern Analysis and Machine Intelligence},
	pages = {7742--7759},
}

@article{scaramuzza_visual_2011,
	title = {Visual {Odometry} [{Tutorial}]},
	volume = {18},
	issn = {1558-223X},
	url = {https://ieeexplore.ieee.org/abstract/document/6096039},
	doi = {10.1109/MRA.2011.943233},
	number = {4},
	urldate = {2024-11-08},
	journal = {IEEE Robotics \& Automation Magazine},
	author = {Scaramuzza, Davide and Fraundorfer, Friedrich},
	month = dec,
	year = {2011},
	note = {Conference Name: IEEE Robotics \& Automation Magazine},
	pages = {80--92},
}

@inproceedings{orchard_efficient_2021,
	address = {Coimbra, Portugal},
	title = {Efficient {Neuromorphic} {Signal} {Processing} with {Loihi} 2},
	isbn = {978-1-66540-144-9},
	url = {https://ieeexplore.ieee.org/document/9605018/},
	doi = {10.1109/SiPS52927.2021.00053},
	urldate = {2023-02-21},
	booktitle = {2021 {IEEE} {Workshop} on {Signal} {Processing} {Systems} ({SiPS})},
	publisher = {IEEE},
	author = {Orchard, Garrick and Frady, E. Paxon and Rubin, Daniel Ben Dayan and Sanborn, Sophia and Shrestha, Sumit Bam and Sommer, Friedrich T. and Davies, Mike},
	month = oct,
	year = {2021},
	pages = {254--259},
}

@inproceedings{livi_current-mode_2009,
	title = {A current-mode conductance-based silicon neuron for address-event neuromorphic systems},
	doi = {10.1109/ISCAS.2009.5118408},
	booktitle = {2009 {IEEE} {International} {Symposium} on {Circuits} and {Systems}},
	publisher = {IEEE},
	author = {Livi, P. and Indiveri, G.},
	year = {2009},
	pages = {2898--2901},
}

@article{lee2020,
	title = {Spike-{FlowNet}: {Event}-{Based} {Optical} {Flow} {Estimation} with {Energy}-{Efficient} {Hybrid} {Neural} {Networks}},
	volume = {12374 LNCS},
	issn = {16113349},
	doi = {10.1007/978-3-030-58526-6_22/COVER},
	urldate = {2023-07-24},
	journal = {Lecture Notes in Computer Science},
	author = {Lee, Chankyu and Kosta, Adarsh Kumar and Zhu, Alex Zihao and Chaney, Kenneth and Daniilidis, Kostas and Roy, Kaushik},
	year = {2020},
	note = {arXiv: 2003.06696
Publisher: Springer Science and Business Media Deutschland GmbH
ISBN: 9783030585259},
	pages = {366--382},
}

@misc{yedutenko_tde-3_2024,
	title = {{TDE}-3: {An} improved prior for optical flow computation in spiking neural networks},
	shorttitle = {{TDE}-3},
	url = {http://arxiv.org/abs/2402.11662},
	doi = {10.48550/arXiv.2402.11662},
	urldate = {2024-09-19},
	publisher = {arXiv},
	author = {Yedutenko, Matthew and Paredes-Valles, Federico and Khacef, Lyes and De Croon, Guido C. H. E.},
	month = feb,
	year = {2024},
	note = {arXiv:2402.11662 [cs]},
}

@inproceedings{lefebvre_mixed-signal_2024,
	title = {A {Mixed}-{Signal} {Near}-{Sensor} {Convolutional} {Imager} {SoC} with {Charge}-{Based} 4b-{Weighted} 5-to-84-{TOPS}/{W} {MAC} {Operations} for {Feature} {Extraction} and {Region}-of-{Interest} {Detection}},
	url = {https://ieeexplore.ieee.org/abstract/document/10528961},
	doi = {10.1109/CICC60959.2024.10528961},
	urldate = {2024-09-19},
	booktitle = {2024 {IEEE} {Custom} {Integrated} {Circuits} {Conference} ({CICC})},
	author = {Lefebvre, Martin and Bol, David},
	month = apr,
	year = {2024},
	note = {ISSN: 2152-3630},
	pages = {1--2},
}

@article{nister_visual_2006,
	title = {Visual odometry for ground vehicle applications},
	volume = {23},
	copyright = {Copyright © 2006 John Wiley \& Sons, Inc.},
	issn = {1556-4967},
	url = {https://onlinelibrary.wiley.com/doi/abs/10.1002/rob.20103},
	doi = {10.1002/rob.20103},
	language = {en},
	number = {1},
	urldate = {2024-08-14},
	journal = {Journal of Field Robotics},
	author = {Nistér, David and Naroditsky, Oleg and Bergen, James},
	year = {2006},
	note = {\_eprint: https://onlinelibrary.wiley.com/doi/pdf/10.1002/rob.20103},
	pages = {3--20},
}

@article{paredes-valles_fully_2024,
	title = {Fully neuromorphic vision and control for autonomous drone flight},
	volume = {9},
	url = {https://www.science.org/doi/abs/10.1126/scirobotics.adi0591},
	doi = {10.1126/scirobotics.adi0591},
	number = {90},
	urldate = {2024-08-05},
	journal = {Science Robotics},
	author = {Paredes-Vallés, F. and Hagenaars, J. J. and Dupeyroux, J. and Stroobants, S. and Xu, Y. and de Croon, G. C. H. E.},
	month = may,
	year = {2024},
	note = {Publisher: American Association for the Advancement of Science},
	pages = {eadi0591},
}

@article{kaufmann_champion-level_2023,
	title = {Champion-level drone racing using deep reinforcement learning},
	volume = {620},
	copyright = {2023 The Author(s)},
	issn = {1476-4687},
	url = {https://www.nature.com/articles/s41586-023-06419-4},
	doi = {10.1038/s41586-023-06419-4},
	language = {en},
	number = {7976},
	urldate = {2024-08-05},
	journal = {Nature},
	author = {Kaufmann, Elia and Bauersfeld, Leonard and Loquercio, Antonio and Müller, Matthias and Koltun, Vladlen and Scaramuzza, Davide},
	month = aug,
	year = {2023},
	note = {Publisher: Nature Publishing Group},
	pages = {982--987},
}

@article{mitrokhin2019,
	title = {Learning sensorimotor control with neuromorphic sensors: {Toward} hyperdimensional active perception},
	volume = {4},
	issn = {24709476},
	doi = {10.1126/scirobotics.aaw6736},
	number = {30},
	journal = {Science Robotics},
	author = {Mitrokhin, A. and Sutor, P. and Fermüller, C. and Aloimonos, Y.},
	month = may,
	year = {2019},
	pmid = {33137724},
	note = {Publisher: American Association for the Advancement of Science},
}

@article{gallego_accurate_2017,
	title = {Accurate {Angular} {Velocity} {Estimation} {With} an {Event} {Camera}},
	volume = {2},
	copyright = {https://ieeexplore.ieee.org/Xplorehelp/downloads/license-information/IEEE.html},
	issn = {2377-3766, 2377-3774},
	url = {http://ieeexplore.ieee.org/document/7805257/},
	doi = {10.1109/LRA.2016.2647639},
	language = {en},
	number = {2},
	urldate = {2024-08-01},
	journal = {IEEE Robotics and Automation Letters},
	author = {Gallego, Guillermo and Scaramuzza, Davide},
	month = apr,
	year = {2017},
	pages = {632--639},
}

@article{gallego_event-based_2018,
	title = {Event-based, 6-{DOF} {Camera} {Tracking} from {Photometric} {Depth} {Maps}},
	volume = {40},
	issn = {0162-8828, 2160-9292, 1939-3539},
	url = {http://arxiv.org/abs/1607.03468},
	doi = {10.1109/TPAMI.2017.2769655},
	language = {en},
	number = {10},
	urldate = {2024-08-01},
	journal = {IEEE Transactions on Pattern Analysis and Machine Intelligence},
	author = {Gallego, Guillermo and Lund, Jon E. A. and Mueggler, Elias and Rebecq, Henri and Delbruck, Tobi and Scaramuzza, Davide},
	month = oct,
	year = {2018},
	note = {arXiv:1607.03468 [cs]},
	pages = {2402--2412},
}

@article{mueggler_continuous-time_2018,
	title = {Continuous-{Time} {Visual}-{Inertial} {Odometry} for {Event} {Cameras}},
	volume = {34},
	issn = {1552-3098, 1941-0468},
	url = {http://arxiv.org/abs/1702.07389},
	doi = {10.1109/TRO.2018.2858287},
	language = {en},
	number = {6},
	urldate = {2024-08-01},
	journal = {IEEE Transactions on Robotics},
	author = {Mueggler, Elias and Gallego, Guillermo and Rebecq, Henri and Scaramuzza, Davide},
	month = dec,
	year = {2018},
	note = {arXiv:1702.07389 [cs]},
	pages = {1425--1440},
}

@inproceedings{zhu_unsupervised_2019,
	title = {Unsupervised {Event}-{Based} {Learning} of {Optical} {Flow}, {Depth}, and {Egomotion}},
	url = {https://openaccess.thecvf.com/content_CVPR_2019/html/Zhu_Unsupervised_Event-Based_Learning_of_Optical_Flow_Depth_and_Egomotion_CVPR_2019_paper.html},
	urldate = {2024-07-31},
	booktitle = {Proceedings of the {IEEE}/{CVF} {Conference} on {Computer} {Vision} and {Pattern} {Recognition}},
	author = {Zhu, Alex Zihao and Yuan, Liangzhe and Chaney, Kenneth and Daniilidis, Kostas},
	year = {2019},
	pages = {989--997},
}

@inproceedings{zhou_unsupervised_2017,
	title = {Unsupervised {Learning} of {Depth} and {Ego}-{Motion} {From} {Video}},
	url = {https://openaccess.thecvf.com/content_cvpr_2017/html/Zhou_Unsupervised_Learning_of_CVPR_2017_paper.html},
	urldate = {2024-07-31},
	booktitle = {Proceedings of the {IEEE} {Conference} on {Computer} {Vision} and {Pattern} {Recognition}},
	author = {Zhou, Tinghui and Brown, Matthew and Snavely, Noah and Lowe, David G.},
	year = {2017},
	pages = {1851--1858},
}

@article{renner_visual_2024,
	title = {Visual odometry with neuromorphic resonator networks},
	volume = {6},
	copyright = {2024 The Author(s), under exclusive licence to Springer Nature Limited},
	issn = {2522-5839},
	url = {https://www.nature.com/articles/s42256-024-00846-2},
	doi = {10.1038/s42256-024-00846-2},
	language = {en},
	number = {6},
	urldate = {2024-07-31},
	journal = {Nature Machine Intelligence},
	author = {Renner, Alpha and Supic, Lazar and Danielescu, Andreea and Indiveri, Giacomo and Frady, E. Paxon and Sommer, Friedrich T. and Sandamirskaya, Yulia},
	month = jun,
	year = {2024},
	note = {Publisher: Nature Publishing Group},
	pages = {653--663},
}

@article{yao_spike-based_2024,
	title = {Spike-based dynamic computing with asynchronous sensing-computing neuromorphic chip},
	volume = {15},
	copyright = {2024 The Author(s)},
	issn = {2041-1723},
	url = {https://www.nature.com/articles/s41467-024-47811-6},
	doi = {10.1038/s41467-024-47811-6},
	language = {en},
	number = {1},
	urldate = {2024-07-31},
	journal = {Nature Communications},
	author = {Yao, Man and Richter, Ole and Zhao, Guangshe and Qiao, Ning and Xing, Yannan and Wang, Dingheng and Hu, Tianxiang and Fang, Wei and Demirci, Tugba and De Marchi, Michele and Deng, Lei and Yan, Tianyi and Nielsen, Carsten and Sheik, Sadique and Wu, Chenxi and Tian, Yonghong and Xu, Bo and Li, Guoqi},
	month = may,
	year = {2024},
	note = {Publisher: Nature Publishing Group},
	pages = {4464},
}

@article{gehrig_low-latency_2024,
	title = {Low-latency automotive vision with event cameras},
	volume = {629},
	copyright = {2024 The Author(s)},
	issn = {1476-4687},
	url = {https://www.nature.com/articles/s41586-024-07409-w},
	doi = {10.1038/s41586-024-07409-w},
	language = {en},
	number = {8014},
	urldate = {2024-07-31},
	journal = {Nature},
	author = {Gehrig, Daniel and Scaramuzza, Davide},
	month = may,
	year = {2024},
	note = {Publisher: Nature Publishing Group},
	pages = {1034--1040},
}

@inproceedings{schoepe_odour_2024,
	title = {Odour {Localization} in {Neuromorphic} {Systems}},
	url = {https://ieeexplore.ieee.org/abstract/document/10558186},
	doi = {10.1109/ISCAS58744.2024.10558186},
	urldate = {2024-07-09},
	booktitle = {2024 {IEEE} {International} {Symposium} on {Circuits} and {Systems} ({ISCAS})},
	author = {Schoepe, Thorben and Drix, Damien and Schüffny, Franz Marcus and Miko, Rebecca and Sutton, Samuel and Chicca, Elisabetta and Schmuker, Michael},
	month = may,
	year = {2024},
	note = {ISSN: 2158-1525},
	pages = {1--5},
}

@article{schoepe_finding_2024,
	title = {Finding the gap: neuromorphic motion-vision in dense environments},
	volume = {15},
	copyright = {2024 The Author(s)},
	issn = {2041-1723},
	shorttitle = {Finding the gap},
	url = {https://www.nature.com/articles/s41467-024-45063-y},
	doi = {10.1038/s41467-024-45063-y},
	language = {en},
	number = {1},
	urldate = {2024-07-09},
	journal = {Nature Communications},
	author = {Schoepe, Thorben and Janotte, Ella and Milde, Moritz B. and Bertrand, Olivier J. N. and Egelhaaf, Martin and Chicca, Elisabetta},
	month = jan,
	year = {2024},
	note = {Publisher: Nature Publishing Group},
	pages = {817},
}

@article{borst2011,
	title = {Seeing {Things} in {Motion}: {Models}, {Circuits}, and {Mechanisms}},
	volume = {71},
	issn = {0896-6273},
	doi = {10.1016/J.NEURON.2011.08.031},
	number = {6},
	urldate = {2024-01-17},
	journal = {Neuron},
	author = {Borst, Alexander and Euler, Thomas},
	month = sep,
	year = {2011},
	pmid = {21943597},
	note = {Publisher: Cell Press},
	pages = {974--994},
}

@article{chicca2014,
	title = {Neuromorphic {Electronic} {Circuits} for {Building} {Autonomous} {Cognitive} {Systems}},
	volume = {102},
	issn = {0018-9219},
	doi = {10.1109/JPROC.2014.2313954},
	number = {9},
	journal = {Proceedings of the IEEE},
	author = {Chicca, Elisabetta and Stefanini, Fabio and Bartolozzi, Chiara and Indiveri, Giacomo},
	month = sep,
	year = {2014},
	pages = {1367--1388},
}

@article{zhu2018,
	title = {{EV}-{FlowNet}: {Self}-{Supervised} {Optical} {Flow} {Estimation} for {Event}-based {Cameras}},
	doi = {10.15607/RSS.2018.XIV.062},
	journal = {Robotics: Science and Systems},
	author = {Zhu, Alex Zihao and Yuan, Liangzhe and Chaney, Kenneth and Daniilidis, Kostas},
	month = feb,
	year = {2018},
	note = {arXiv: 1802.06898v4
Publisher: MIT Press Journals},
}

@article{mitrokhin2018,
	title = {Event-{Based} {Moving} {Object} {Detection} and {Tracking}},
	issn = {21530866},
	doi = {10.1109/IROS.2018.8593805},
	urldate = {2023-07-24},
	journal = {IEEE International Conference on Intelligent Robots and Systems},
	author = {Mitrokhin, Anton and Fermuller, Cornelia and Parameshwara, Chethan and Aloimonos, Yiannis},
	month = dec,
	year = {2018},
	note = {arXiv: 1803.04523
Publisher: Institute of Electrical and Electronics Engineers Inc.
ISBN: 9781538680940},
	pages = {6895--6902},
}

@article{vitale2021,
	title = {Event-driven {Vision} and {Control} for {UAVs} on a {Neuromorphic} {Chip}},
	volume = {2021-May},
	issn = {10504729},
	doi = {10.1109/ICRA48506.2021.9560881},
	urldate = {2023-07-24},
	journal = {Proceedings - IEEE International Conference on Robotics and Automation},
	author = {Vitale, Antonio and Renner, Alpha and Nauer, Celine and Scaramuzza, Davide and Sandamirskaya, Yulia},
	year = {2021},
	note = {arXiv: 2108.03694
Publisher: Institute of Electrical and Electronics Engineers Inc.
ISBN: 9781728190778},
	pages = {103--109},
}

@article{censi2014,
	title = {Low-latency event-based visual odometry},
	issn = {10504729},
	doi = {10.1109/ICRA.2014.6906931},
	urldate = {2023-07-24},
	journal = {Proceedings - IEEE International Conference on Robotics and Automation},
	author = {Censi, Andrea and Scaramuzza, Davide},
	month = sep,
	year = {2014},
	note = {Publisher: Institute of Electrical and Electronics Engineers Inc.
ISBN: 9781479936854},
	pages = {703--710},
}

@article{cuadrado2023,
	title = {Optical flow estimation from event-based cameras and spiking neural networks},
	volume = {17},
	issn = {1662453X},
	doi = {10.3389/FNINS.2023.1160034/BIBTEX},
	urldate = {2023-07-24},
	journal = {Frontiers in Neuroscience},
	author = {Cuadrado, Javier and Rançon, Ulysse and Cottereau, Benoit R. and Barranco, Francisco and Masquelier, Timothée},
	month = may,
	year = {2023},
	note = {arXiv: 2302.06492
Publisher: Frontiers Media S.A.},
	pages = {1160034},
}

@article{nunes2022,
	title = {Robust {Event}-{Based} {Vision} {Model} {Estimation} by {Dispersion} {Minimisation}},
	volume = {44},
	issn = {19393539},
	doi = {10.1109/TPAMI.2021.3130049},
	number = {12},
	urldate = {2023-06-26},
	journal = {IEEE Transactions on Pattern Analysis and Machine Intelligence},
	author = {Nunes, Urbano Miguel and Demiris, Yiannis},
	month = dec,
	year = {2022},
	pmid = {34813470},
	note = {Publisher: IEEE Computer Society},
	pages = {9561--9573},
}

@article{posch2011,
	title = {A {QVGA} 143 {dB} dynamic range frame-free {PWM} image sensor with lossless pixel-level video compression and time-domain {CDS}},
	volume = {46},
	issn = {00189200},
	doi = {10.1109/JSSC.2010.2085952},
	number = {1},
	urldate = {2023-07-20},
	journal = {IEEE Journal of Solid-State Circuits},
	author = {Posch, Christoph and Matolin, Daniel and Wohlgenannt, Rainer},
	month = jan,
	year = {2011},
	pages = {259--275},
}

@article{mead1988,
	title = {A silicon model of early visual processing},
	volume = {1},
	issn = {0893-6080},
	doi = {10.1016/0893-6080(88)90024-X},
	number = {1},
	urldate = {2023-07-20},
	journal = {Neural Networks},
	author = {Mead, Carver A. and Mahowald, M. A.},
	month = jan,
	year = {1988},
	note = {Publisher: Pergamon},
	pages = {91--97},
}

@article{mahowald1994,
	title = {The {Silicon} {Retina}},
	doi = {10.1007/978-1-4615-2724-4_2},
	urldate = {2023-07-19},
	journal = {An Analog VLSI System for Stereoscopic Vision},
	author = {Mahowald, Misha},
	year = {1994},
	note = {Publisher: Springer, Boston, MA},
	pages = {4--65},
}

@article{gehrig_e-raft_2021,
	title = {E-{RAFT}: {Dense} {Optical} {Flow} from {Event} {Cameras}},
	doi = {10.1109/3DV53792.2021.00030},
	urldate = {2023-06-23},
	journal = {Proceedings - 2021 International Conference on 3D Vision, 3DV 2021},
	author = {Gehrig, Mathias and Millhausler, Mario and Gehrig, Daniel and Scaramuzza, Davide},
	year = {2021},
	note = {arXiv: 2108.10552
Publisher: Institute of Electrical and Electronics Engineers Inc.
ISBN: 9781665426886},
	pages = {197--206},
}

@article{ye20,
	title = {Unsupervised learning of dense optical flow, depth and egomotion with event-based sensors},
	issn = {21530866},
	doi = {10.1109/IROS45743.2020.9341224},
	urldate = {2023-04-14},
	journal = {IEEE International Conference on Intelligent Robots and Systems},
	author = {Ye, Chengxi and Mitrokhin, Anton and Fermuller, Cornelia and Yorke, James A. and Aloimonos, Yiannis},
	month = oct,
	year = {2020},
	note = {Publisher: Institute of Electrical and Electronics Engineers Inc.
ISBN: 9781728162126},
	pages = {5831--5838},
}

@article{bartolozzi2007,
	title = {Synaptic {Dynamics} in {Analog} {VLSI}},
	volume = {19},
	issn = {0899-7667},
	doi = {10.1162/neco.2007.19.10.2581},
	number = {10},
	journal = {Neural Computation},
	author = {Bartolozzi, Chiara and Indiveri, Giacomo},
	month = oct,
	year = {2007},
	pages = {2581--2603},
}

@article{milde2018,
	title = {Spiking {Elementary} {Motion} {Detector} in {Neuromorphic} {Systems}},
	volume = {30},
	issn = {0899-7667},
	doi = {10.1162/neco_a_01112},
	number = {9},
	journal = {Neural Computation},
	author = {Milde, M. B. and Bertrand, O. J. N. and Ramachandran, H. and Egelhaaf, M. and Chicca, E.},
	month = sep,
	year = {2018},
	pages = {2384--2417},
}


\end{refsection}

\end{document}